\pdfoutput=1
\documentclass[10pt,twocolumn,letterpaper]{article}

\usepackage{iccv}
\usepackage{times}
\usepackage{epsfig}
\usepackage{graphicx}
\usepackage{amsmath}
\usepackage{amssymb}
\usepackage{enumitem}

\usepackage[usestackEOL]{stackengine}
\usepackage{wrapfig}    
\usepackage{diagbox}    
\usepackage{multirow}
\usepackage{booktabs}
\usepackage{bm}
\usepackage[table]{xcolor}
\usepackage{algorithm}
\usepackage{algorithmic}
\definecolor{mygray}{gray}{0.9}

\usepackage{pifont}

\usepackage[final]{pdfpages}

\usepackage[pagebackref=true,breaklinks=true,letterpaper=true,colorlinks,bookmarks=false]{hyperref}

\iccvfinalcopy 

\def\iccvPaperID{2534} 

\newcommand{\egs}{\textit{e.g.}}
\newcommand{\ies}{\textit{i.e.}}
\newcommand\independent{\protect\mathpalette{\protect\independenT}{\perp}}
\def\independenT#1#2{\mathrel{\rlap{$#1#2$}\mkern2mu{#1#2}}}

\usepackage[table]{xcolor}
\definecolor{mygray}{gray}{0.9}

\begin{document}

\title{Causal Attention for Unbiased Visual Recognition}
\author{Tan Wang$^
{1}$, 
\quad Chang Zhou$^{2}$, 
\quad Qianru Sun$^{3}$,
\quad Hanwang Zhang$^{1}$\\
\small $^1$Nanyang Technological University~~
\small $^2$Damo Academy, Alibaba Group~~
\small $^3$Singapore Management University\\
\small {\texttt{TAN317@e.ntu.edu.sg}},~
\small {\texttt{zhouchang.zc@alibaba-inc.com}},~
\small {\texttt{qianrusun@smu.edu.sg}},~
\small {\texttt{hanwangzhang@ntu.edu.sg}}
\\
}

\maketitle

\begin{abstract}

Attention module does not always help deep models learn \emph{causal features} that are robust in any confounding context, e.g., a foreground object feature is invariant to different backgrounds. This is because the confounders trick the attention to capture spurious correlations that benefit the prediction when the training and testing data are IID (identical \& independent distribution); while harm the prediction when the data are OOD (out-of-distribution). The sole fundamental solution to learn causal attention is by causal intervention, which requires additional annotations of the confounders, e.g., a ``dog'' model is learned within  ``grass+dog'' and ``road+dog'' respectively, so the ``grass'' and ``road'' contexts will no longer confound the ``dog'' recognition.  However, such annotation is not only prohibitively expensive, but also inherently problematic, as the confounders are elusive in nature. In this paper, we propose a causal attention module (CaaM) that  self-annotates the confounders in unsupervised fashion. In particular, multiple CaaMs can be stacked and integrated in conventional attention CNN and self-attention Vision Transformer. In OOD settings, deep models with CaaM outperform those without it significantly; even in IID settings, the attention localization is also improved by CaaM, showing a great potential in applications that require robust visual saliency. Codes are available at {\small\url{https://github.com/Wangt-CN/CaaM}}.

\end{abstract}


\section{Introduction}
\label{sec:intro}
\begin{figure}[t]
\begin{center}
\includegraphics[width=0.45\textwidth]{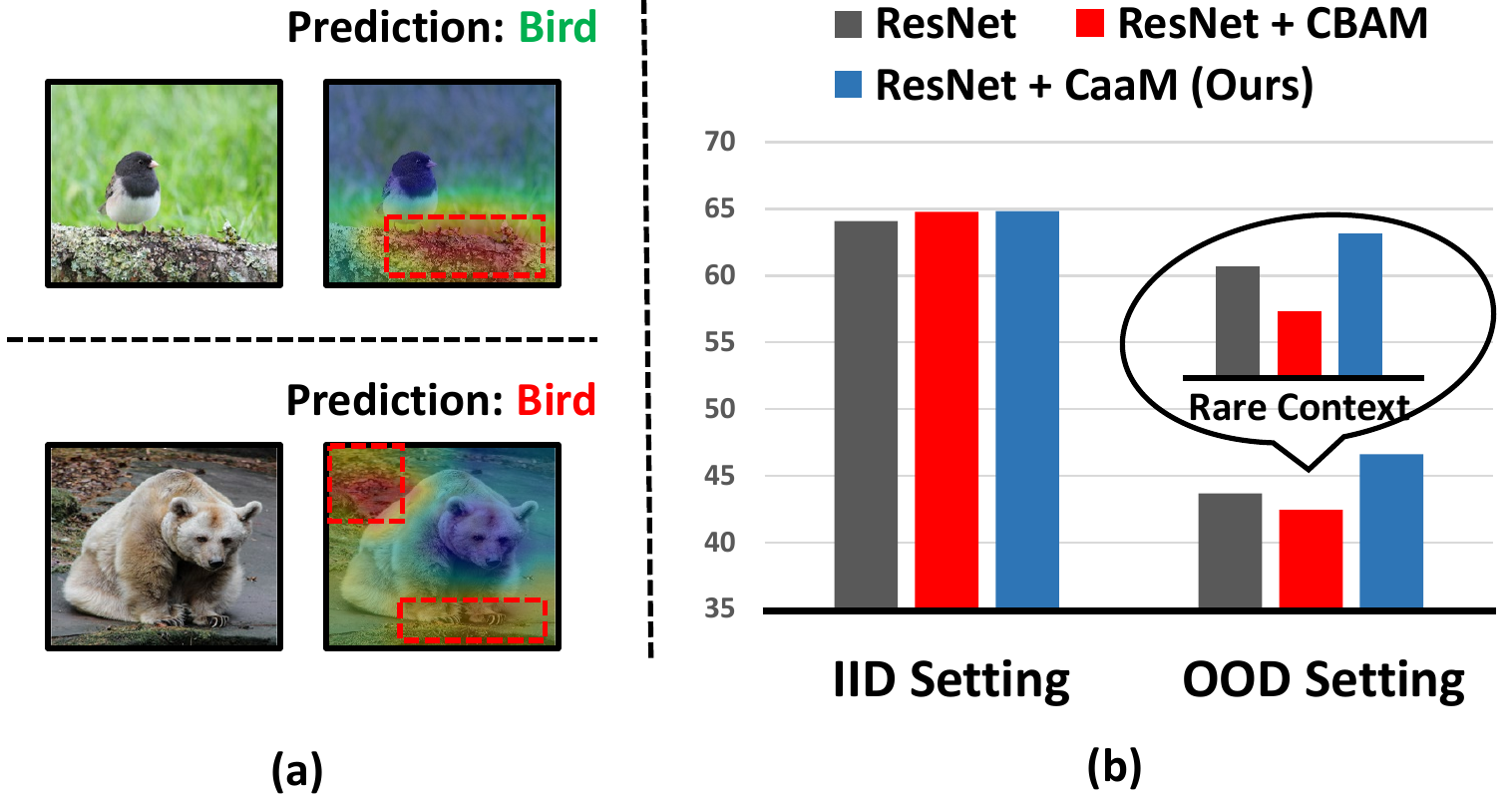}
\end{center}
  \caption{(a) The qualitative attention maps of two images in NICO~\cite{he2021towards} using ResNet18 with CBAM~\cite{woo2018cbam}. (b) The accuracies of three methods: ResNet18,  ResNet18+CBAM~\cite{woo2018cbam} and + CaaM (Ours) in both IID and OOD settings.
  }
\vspace{-0.2cm}
\label{fig:figure1}
\end{figure}

Do you think attention~\cite{xu2015show,vaswani2017attention} would always capture the salient regions in an image? No, as shown in Figure~\ref{fig:figure1} (a, top), due to the lack of region-level labels, ``learning to attend'' is a \emph{de facto} weakly-supervised task. Or do you think attention would always  improve  performance? Probably yes, after all, ``Attention is All You Need''~\cite{devlin2018bert, dosovitskiy2020image, ramesh2021zero}. As shown in Figure~\ref{fig:figure1} (a, top), even if the attended region is wrong, the model still makes correct predictions. In conventional IID settings, where the training and testing data are identically and independently distributed, the model equipped with attention is indeed better (red bar is higher than black bar in Figure~\ref{fig:figure1} (b)).

However,  few people realize that attention may do evil in OOD settings, where the testing data are out of the training distribution. For example, as shown in Figure~\ref{fig:figure1} (a, top), the attention considers the ``ground'' region as the visual cue of the ``bird'' class, because most training ``bird'' images are in ``ground'' context; but, when the test image is ``bear in ground'' (bottom), the attention misleads the model to still predict ``bird''.  Figure~\ref{fig:figure1} (b) reports that the attention model is even worse than the non-attention baseline in OOD setting (red bar is lower than black bar), where the gap is further amplified by the rare object and context combination in training.
Unfortunately, when we deploy such vision systems in critical domains such as car autopilot, it is often the rare case that causes fatal accidents, \eg, recognizing a ``white'' truck as ``white'' clouds\footnote[1]{\url{https://www.youtube.com/watch?v=X3hrKnv0dPQ}}.

Astute readers who are knowledgeable in causality~\cite{holland1986statistics, pearl2009causality} may point out that the key reason for the bipolar role of attention in IID and OOD is due to the confounding effect~\cite{wang2020visual,zhang2020causal,yue2020interventional,yang2020deconfounded}. In visual recognition, the causal pursuit between the input image $X$ and the output label $Y$ is confounded by a common cause: the context $S$. To see the effect, during data collection, $X$ is usually found in $S$, and thus $S$ is a contextual cue to recognize $Y$ (\ie, $X\leftarrow S\rightarrow Y$). After training, the model recklessly exploits the statistical cues of $S$ as a feature of $X$ to predict  $Y$ (\eg, most training ``bird'' is on ``ground'' in Figure~\ref{fig:figure1} (a, top)); however, in testing, if $X\not\rightarrow Y$ (\eg, ``bear'' image $\not\rightarrow$ ``bird''), seeing $S$ misleads $X\rightarrow Y$ (\eg, the ``ground'' pattern always recalls ``bird''). In Section~\ref{sec:preliminary}, we will revisit the above causality in detail.

\begin{figure}[t]
\begin{center}
\includegraphics[width=0.48\textwidth]{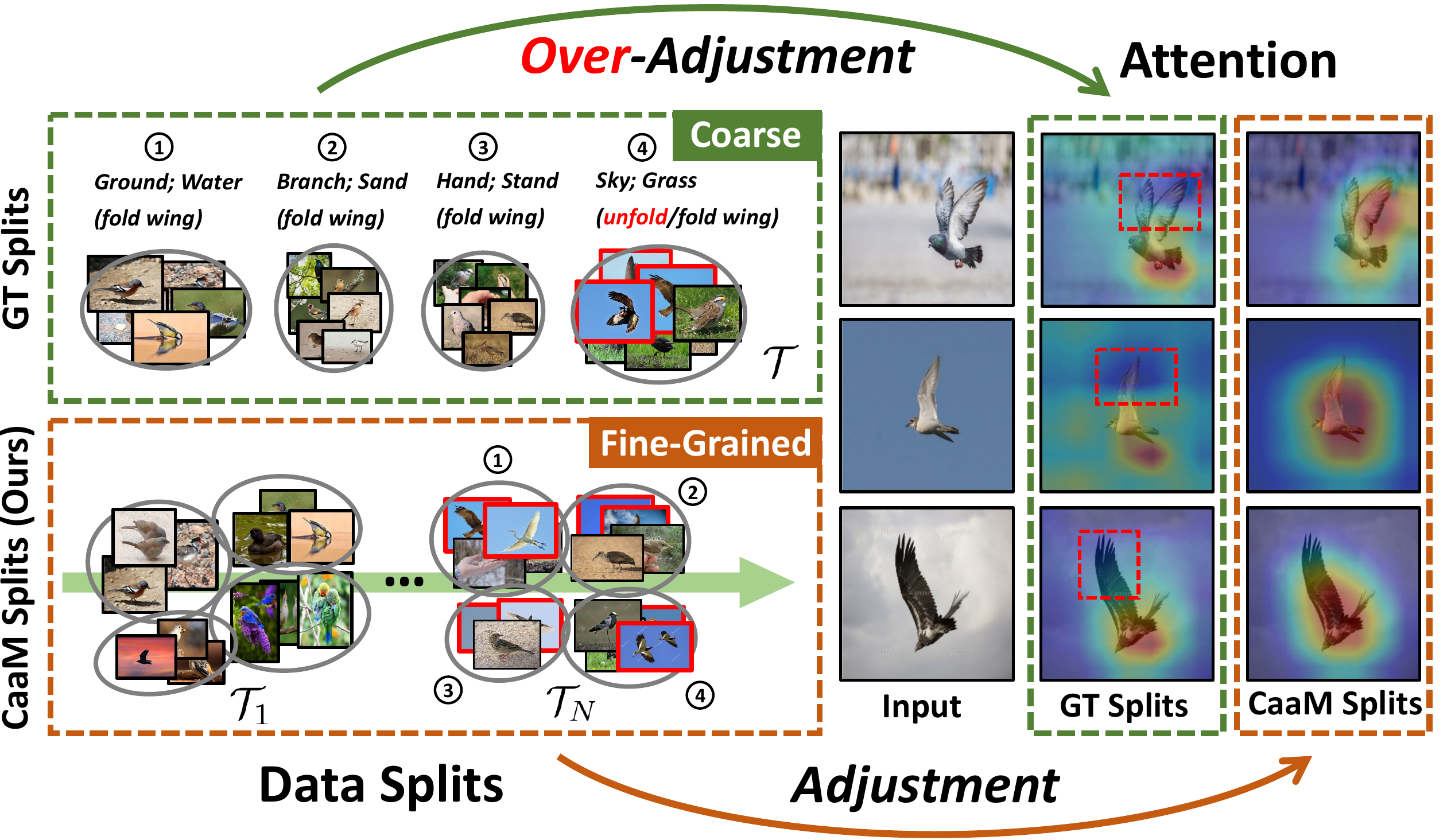}
\end{center}
  \caption{The comparison of attention maps between the partition-based intervention methods~\cite{arjovsky2019invariant,teney2020unshuffling} and our unsupervised CaaM with each training data splits.
  $\mathcal{T}_N$ denotes the $N$-th partition. Birds unfolding wings are highlighted with red boxes.
}
\vspace{-0.4cm}
\label{fig:figure-overadjust}
\end{figure}

The sole solution to mitigate the confounding bias is by \emph{causal intervention}~\cite{pearl2009causality}. For example,  Arjovsky~\etal~\cite{arjovsky2019invariant} and Teney~\etal~\cite{teney2020unshuffling} suggest to collect ``bird'' images under every context (\ie, adjusting the contexts of ``bird''). In each context split (\eg, ``ground'', ``water'' and ``sky''), the attention does not attend to the context as it is no longer discriminative in the split. So, the combined ``ground/water/sky + bird'' attention will tend to focus on the ``bird'' unbiased towards any context. However, it is  impractical to perform the causal intervention like above. Despite the expensive cost of extra annotations, it suffers from the following deficiency.

In practice, it is impossible to collect the samples of a class in any context, \eg, it is hard to find ``fish'' in ``sky''. Technically, such absent context of a class violates the confounder positivity assumption of casual intervention~\cite{hernan2010causal} (see Section~\ref{sec:ablation_study} for the poor performance caused by the violation). Therefore, we have to merge the ground-truth contexts into bigger splits to include all classes (\eg, merging ``water'' and ``ground'' as one split in Figure~\ref{fig:figure-overadjust} (left top)).

However, such coarser contexts will lead to the \emph{over-adjustment} problem --- the intervention not only removes the context, but also hurts beneficial causalities (\eg, the object part). Figure~\ref{fig:figure-overadjust} (left top) illustrates a real example. Recall that the aforementioned context split-based intervention removes the non-causal features of different contexts. Unfortunately, the causal feature of ``bird''---``wing''---is also removed (see red dashed boxes in the attention). This is because all the birds in ``sky'' context unfold their wings: split \ding{175} does not only represent ``sky'' and ``grass'', but also ``wing''. We formally formulate this problem in \emph{improper causal intervention} in Section~\ref{sec:preliminary}.

In this paper, we propose a causal attention module (CaaM $\backslash$ka:m$\backslash$) which generates data partition iteratively and self-annotates the confounders progressively to overcome the over-adjustment problem.
Compared to the coarser contexts, multiple CaaM partitions are fine-grained and more exact to describe the comprehensive confounder.
As shown in Figure~\ref{fig:figure-overadjust} (left bottom), each split of partition $\mathcal{T}_N$ has images of ``bird'' unfolding ``wings'' (see images in red boxes).
This encourages the model to capture the ``wing'' feature (see the improved visual attention), because ``wing'' no longer co-varies with the \ding{175} ``Sky; Grass'' context. Technically, besides a standard attention that attends to the desired causal features (\eg, foreground), CaaM has a complementary attention that deliberately captures the confounding effect (\eg, background). The two disentangled attentions are optimized in an adversarial minimax fashion, which progressively constitute the confounder set and mitigates the confounding bias in unsupervised fashion.

We analyze how CaaM learns better causal features than existing baselines in Section~\ref{sec:caam}. 
In Section~\ref{sec:implementation}, we show two deployment examples on the popular attention-based deep models: CBAM-based CNN~\cite{woo2018cbam} and Transformer-based T2T-ViT~\cite{yuan2021tokens}. Extensive qualitative and quantitative experimental results in Section~\ref{sec:exp} demonstrate the consistent gain achieved by CaaM. 

Our technical contributions are summarized as:
\vspace{-2mm}
\begin{itemize}[leftmargin=1.2em]
  \setlength\itemsep{0em}
  \item A novel yet practical visual attention module CaaM who learns causal features that are robust in OOD settings without sacrificing the performance in IID settings.
  \item We offer a causality-theoretic analysis to guarantee the superiority of CaaM.
  \item  The design of CaaM is generic to popular deep networks.
\end{itemize}

\section{Related Work}
\label{sec:related_work}

\noindent\textbf{Visual Attention}. We consider both conventional attention~\cite{woo2018cbam,hu2018squeeze} and the recent self-attention~\cite{vaswani2017attention, dosovitskiy2020image, touvron2020training, yuan2021tokens, heo2021rethinking, vaswani2021scaling, wang2018non}. Over the past years, although they had evolved into various models, the key mechanism is still to select the informative features (subject to a context or token query)~\cite{corbetta2002control,mnih2014recurrent,chen2017sca}. Due to that the selection has no localized strong supervision, attention is inherently biased in OOD settings. Most recently, Yang~\etal~\cite{yang2021causal} also investigated the biased attention. However, our CaaM is fundamentally different: 
1) Different assumptions. \cite{yang2021causal} is for visual-language tasks and assumes the mediator is visible from the vision-language context; however, in general visual recognition, this requirement is inapplicable. 
2) Different methods. \cite{yang2021causal} uses front-door adjustment~\cite{pearl2016causal}, while our CaaM is back-door adjustment. More importantly, CaaM can self-annotate the confounder in an \textit{unsupervised} way. In this view, CaaM is also technically different from recent visual causal inference works~\cite{wang2020visual, zhang2020causal, tang2020long, niu2020counterfactual,yue2021counterfactual,hu2021distilling}.

\noindent\textbf{OOD Generalization.} Machine learning is always challenged by OOD problems~\cite{liang2017enhancing,hendrycks2016baseline,achille2018information}, such as debiasing~\cite{geirhos2018imagenet,kim2019learning,huang2017arbitrary,clark2019don,wang2019learning,li2019repair}, domain adaption~\cite{ben2007analysis, muandet2013domain, ganin2016domain, tzeng2017adversarial, gong2016domain} and long-tailed recognition~\cite{khan2017cost,mahajan2018exploring,shen2016relay}. We focus on the most challenging yet practical OOD setting~\cite{bahng2020learning, he2021towards, hendrycks2019natural} where the OOD visual semantics are unlabeled (different from long-tailed) and ubiquitous (different from domain adaptation). 
Moreover, we follow and reveal the recent progress of invariant risk minimization (IRM)~\cite{arjovsky2019invariant,krueger2020out, rosenfeld2020risks,creager2020environment,ahuja2020empirical, ahuja2020invariant,liu2021heterogeneous,ye2021adversarial} as a kind of causal intervention, which however suffers from the over-adjustment discussed later. 
Our CaaM utilizes the complementary attention and an iterative adversarial training pipeline to overcome this problem.

\section{CaaM: Causal Attention Module}
\label{sec:caam}

\subsection{Causal Preliminaries}
\label{sec:preliminary}

\begin{figure}[h]
\begin{center}
\includegraphics[width=0.45\textwidth]{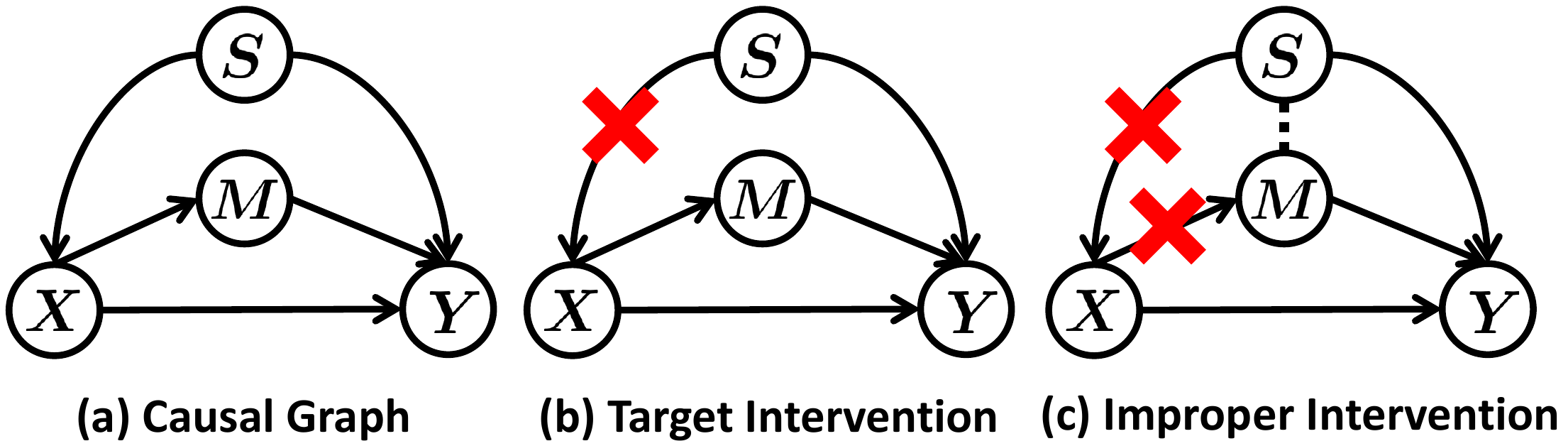}
\vspace{-0.3cm}
\end{center}
  \caption{The causal graphs of visual recognition.}
\vspace{-0.3cm}
\label{fig:causal_graph}
\end{figure}

\noindent\textbf{Causal View of Biased Recognition.}
We introduce the formulation of causality for visual recognition tasks by using a Structure Causal Model (SCM)~\cite{pearl2009causality}. We build this SCM by inspecting on the causalities among the key components: image $X$, label $Y$, mediator $M$, and confounder $S$.
We illustrate the SCM in Figure~\ref{fig:causal_graph} (a) where each direct link denotes a causal relationship between two nodes.

\noindent\bm{$X \rightarrow Y$} denotes the desired causal effect from image content $X$ to label $Y$, as image is labeled for its content. We call a recognition model is unbiased if it identifies $X \rightarrow Y$.

\noindent\bm{$X \leftarrow S \rightarrow Y$}. $S \rightarrow X$ denotes that unstable context $S$ determines what to picture in image $X$~\cite{zhang2020causal}. 
For instance, $S$ determines where to put ``birds'' and ``ground'' in an image. 
$S \rightarrow Y$ exists because model inevitablely uses the contextual cue to recognize $Y$.
In the SCM, we can clearly see how $S$ confounds $X$ and $Y$ via the backdoor path $X \leftarrow S \rightarrow Y$. Taking the bear-bird example again (Figure~\ref{fig:figure1} (a)), though the ``bear'' image ($X$) has no causal relationship with the label of ``bird'' ($Y$), the backdoor path creates a spurious correlation between them (through $S$) and thus yields the wrong prediction of ``bird'' from a ``bear'' image.

\noindent\bm{$X \rightarrow M \rightarrow Y$} is a beneficial causal effect for robust recognition, where $M$ is a mediator that are invariant in different distributions. 
For example, $M$ can be discriminative object parts, \eg, ``bird'' has ``wing''.
Note that $M$ can be hidden in the causal path $X\rightarrow Y$. Here we define it as an explicit graph node for the convenience of following mathematical derivations.

\begin{figure}[t]
\begin{center}
\includegraphics[width=0.46\textwidth]{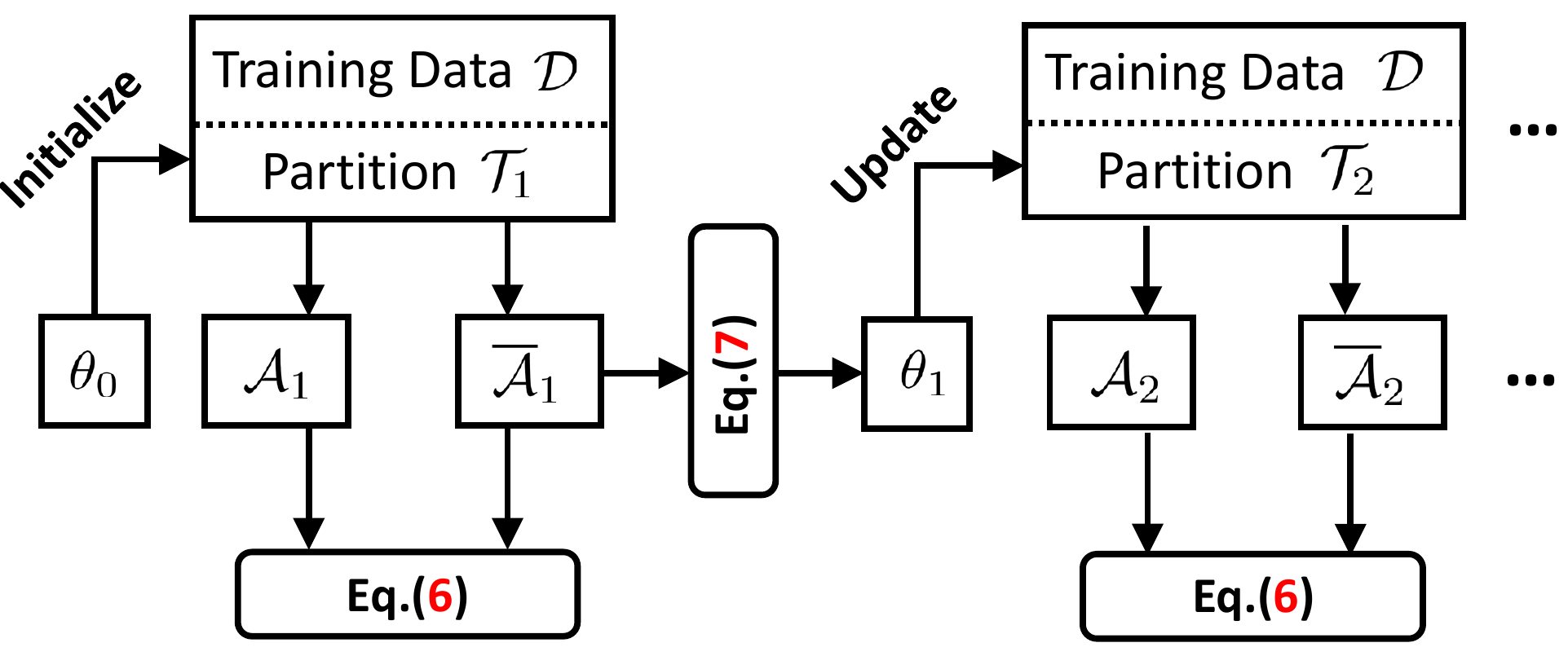}
\vspace{-0.3cm}
\end{center}
  \caption{The training pipeline of our CaaM. In each iteration, it contains a Mini-Game: Joint Training (Eq.~\eqref{eq:jointloss}) and a Maxi-Game: Partition Update (Eq.~\eqref{eq:maxigame}).
  The subscript of $\mathcal{T}, \mathcal{A}$, and $\theta$  is iteration index. The resultant attention for unbiased recognition is $\mathcal{A}_N$ after $N$ iterations.}
\vspace{-0.4cm}
\label{fig:pipeline}
\end{figure}

\noindent\textbf{Causal Intervention by Data Partition.}
Data partition (or environment split)~\cite{arjovsky2019invariant} is an effective implementation of causal intervention. It first partitions the training data into a set of hard splits $\mathcal{T}=\{t_1,...,t_m\}$, each of which represents a confounder stratum, allowing the model trained across different splits invariant to the confounder. We show that data partition is equivalent to the well-known backdoor adjustment~\cite{pearl2009causality}:
\begin{equation}
    P(Y|do(X)) = \sum_{t\in \mathcal{T}} P(Y|X,t)P(t),
    \label{eq:backdoor}
\end{equation}
where $P(Y|X, t)$ denotes the prediction of the classifier trained
in split $t$ and $P(t):=1/m$.
We illustrate $P(Y|do(X))$ in 
Figure~\ref{fig:causal_graph} (b).
The interpretation is that $do(X)$ cuts off the confounding path $X \leftarrow S \rightarrow Y$, leaving only robust paths $X \rightarrow Y$ and $X \rightarrow M \rightarrow Y$. However, existing methods based on data partition~\cite{arjovsky2019invariant, teney2020unshuffling} only assumes a single yet small set of splits, which is far from sufficient for Eq.~\eqref{eq:backdoor}.

\noindent\textbf{Improper Causal Intervention.}
In visual recognition, a perfect partition as in Eq.~\eqref{eq:backdoor} is not easy to obtain because the conventional context-based partition annotation~\cite{arjovsky2019invariant, he2021towards} does not disentangle the confounder and mediator. Thus, straightforwardly adjusting the mediator hurts feature learning~\cite{tang2020long}. Below, we elaborate the reasons using causal formulations.
Suppose that the data partition $\mathcal{T}$ only contains the confounder. Then, we can mitigate $S$ without blocking $M$. By applying Bayes rules, Eq.~\eqref{eq:backdoor} can be re-written as:
{\small
\begin{equation}
    P(Y|do(X)) = \sum_{s\in S} \sum_{m\in M} P(Y|X, s, m)\underline{P(m|X)}P(s).
    \label{eq:true_effect}
\end{equation}}
However, if each split in $\mathcal{T}$ contains both $S$ and $M$, \ies, $(S \not\independent M) \mid X$.
Eq.~\eqref{eq:backdoor} will be re-derived to the false effect estimation:
{\small
\begin{equation}
    P(Y|do(X)) = \sum_{s\in S} \sum_{m\in M} P(Y|X, s, m)\underline{P(m|X, s)}P(s),
    \label{eq:false_effect}
\end{equation}}
where $P(m|X, s)$ is not equal to $P(m|X)$ in Eq.~\eqref{eq:true_effect} due to $(S \not\independent M) \mid X$. This means that the improper partition $\mathcal{T}$ indeed cuts off the robust mediation effect of $X\rightarrow M\rightarrow Y$, as shown in Figure~\ref{fig:causal_graph} (c).

\subsection{Training Pipeline}
\label{sec:caam}

The iterative training pipeline of any model equipped with CaaM is illustrated in Figure~\ref{fig:pipeline}. To enlarge the split number in Eq.~\eqref{eq:backdoor}, we discover partition $\mathcal{T}_i$ in each step. After $N$ steps training, we can approximate Eq.~\eqref{eq:backdoor} by $P(Y|do(X)) \approx \sum\nolimits^N_i\sum\nolimits_{t\in\mathcal{T}_i}P(Y|X,t)P(t)$. For the disentanglement of confounder and mediator, we design a pair of complementary attention modules $\mathcal{A}$ and $\overline{\mathcal{A}}$, where $\mathcal{A}$ is for attending to features of causal effect $X\rightarrow M\rightarrow Y$ and $X\rightarrow Y$, while $\overline{\mathcal{A}}$ is for attending to the confounding effect $X\leftarrow S\rightarrow Y$. Note that the roles of $\mathcal{A}$ and $\overline{\mathcal{A}}$  are adversarial, as the former aims to predict correctly using robust feature while the latter aims to capture bias. Therefore, the adversarial training encourages disentanglement, and we can use $\overline{\mathcal{A}}$ to update the partition $\mathcal{T}_{i+1}$. We illustrate the convergence of our training pipeline in Appendix. Next, we will detail the training losses.

\noindent\textbf{Cross-Entropy Loss}. This loss is to ensure that $\mathcal{A}$ and $\overline{\mathcal{A}}$ combined will capture the biased total effect from $X\rightarrow Y$ regardless of causal or confounding effects; otherwise, they may disrespect the training data generative causality as assumed in Figure~\ref{fig:causal_graph} (a). Note that such biased training practice is widely adopted in unbiased models~\cite{cadene2019rubi,niu2020counterfactual,tang2020unbiased}. 
\begin{equation}
    \mathrm{XE}(f, \widetilde{x}, \mathcal{D}) = \mathbb{E}_{(x,y)\in \mathcal{D}}~\ell\left(f (\widetilde{x}), y\right),
    \label{eq:celoss}
\end{equation}
where $\widetilde{x}=\mathcal{A}(x)\circ\overline{\mathcal{A}}(x)$ and $\circ$ denotes feature addition, $f$ is a linear classifier,  and $\ell$ is the cross-entropy loss function.

\noindent\textbf{Invariant Loss}~\cite{arjovsky2019invariant}. This loss is for learning $\mathcal{A}$ that is split-invariant made by the causal intervention in Eq.~\eqref{eq:backdoor} with incomplete confounder partition $\mathcal{T}_i$:
\begin{equation}
\begin{split}
    \mathrm{IL}(g, \mathcal{A}(x), \mathcal{T}_i)=&\sum_{t\in \mathcal{T}_i} \mathrm{XE}(g, \mathcal{A}(x), t) \\
    &+\lambda \| \nabla_{\mathbf{w}=\mathbf{1.0}} \mathrm{XE}(\mathbf{w}, \mathcal{A}(x), t) \|_2^2,
    \label{eq:illoss}
\end{split}
\end{equation}
where $t$ is a data split, $g$ is a linear classifier for robust prediction, $\mathbf{w}$ stands for a dummy classifier~\cite{arjovsky2019invariant} used to calculate gradient penalty across splits and $\lambda$ is the weight. 
During inference, $g(\mathcal{A}(x))$ will be deployed for unbiased recognition.
See Appendix for further details.

\noindent\textbf{Adversarial Training}. This training disentangles $\mathcal{A}$ and $\overline{\mathcal{A}}$ with a Mini-Game and a Maxi-Game.
Intuitively, the Maxi-Game takes the confounder feature in $\overline{\mathcal{A}}(x)$ to generate the data partition $\mathcal{T}_{i}$ (causal feature does not contribute to maximization). While the Mini-Game exclude such confounder feature from $\mathcal{A}(x)$ with $\mathcal{T}_{i}$ (confounder feature does not contribute to minimization).

\noindent\textit{Mini-Game:}
It is a joint training with XE and IL, plus a new adversary classifier $h$ that specializes in the confounding effect caused by $\overline{\mathcal{A}}(x)$:
{\footnotesize
\begin{equation}
    \mathop{\min}_{\mathcal{A}, \overline{\mathcal{A}}, f, g, h} \mathrm{XE}(f, \widetilde{x}, \mathcal{D}) + \mathrm{IL}(g, \mathcal{A}(x),\mathcal{T}_i) + \mathrm{XE}(h, \overline{\mathcal{A}}(x), \mathcal{D}),
    \label{eq:jointloss}
\end{equation}}
\noindent\textit{Maxi-Game:} 
A good partition update should captures stronger confounder that is NOT split invariant:

{\small
\begin{equation}
    \mathop{\max}_{\theta}~\mathrm{IL}(h, \overline{\mathcal{A}}(x), \mathcal{T}_i(\theta))
    \label{eq:maxigame}
\end{equation}}
where $\mathcal{T}_i(\theta)$ denotes partition $\mathcal{T}_i$ is decided by $\theta\in\mathbb{R}^{K\times m}$, $K$ is the total number of training samples and $m$ is the number of splits in a partition. $\theta_{p,q}$ is the probability of the $p$-th sample belonging to the $q$-th split ($t_q \in \mathcal{T}_i$).


\subsection{Implementations of CaaM}
\label{sec:implementation}

We implement the proposed CaaM on two popular attention-based deep models:  CBAM-based CNN~\cite{woo2018cbam} and Transformer-based T2T-ViT~\cite{yuan2021tokens}. We call the result models as CNN-CaaM and ViT-CaaM, respectively.
For simplicity, in this section we use $\mathbf{c}$ and $\mathbf{s}$ to denote the causal and confounder feature (\ies, $\mathbf{c}=\mathcal{A}(x)$ and $\mathbf{s}=\overline{\mathcal{A}}(x)$).
\begin{figure*}[t]
\begin{center}
\includegraphics[width=0.9\textwidth]{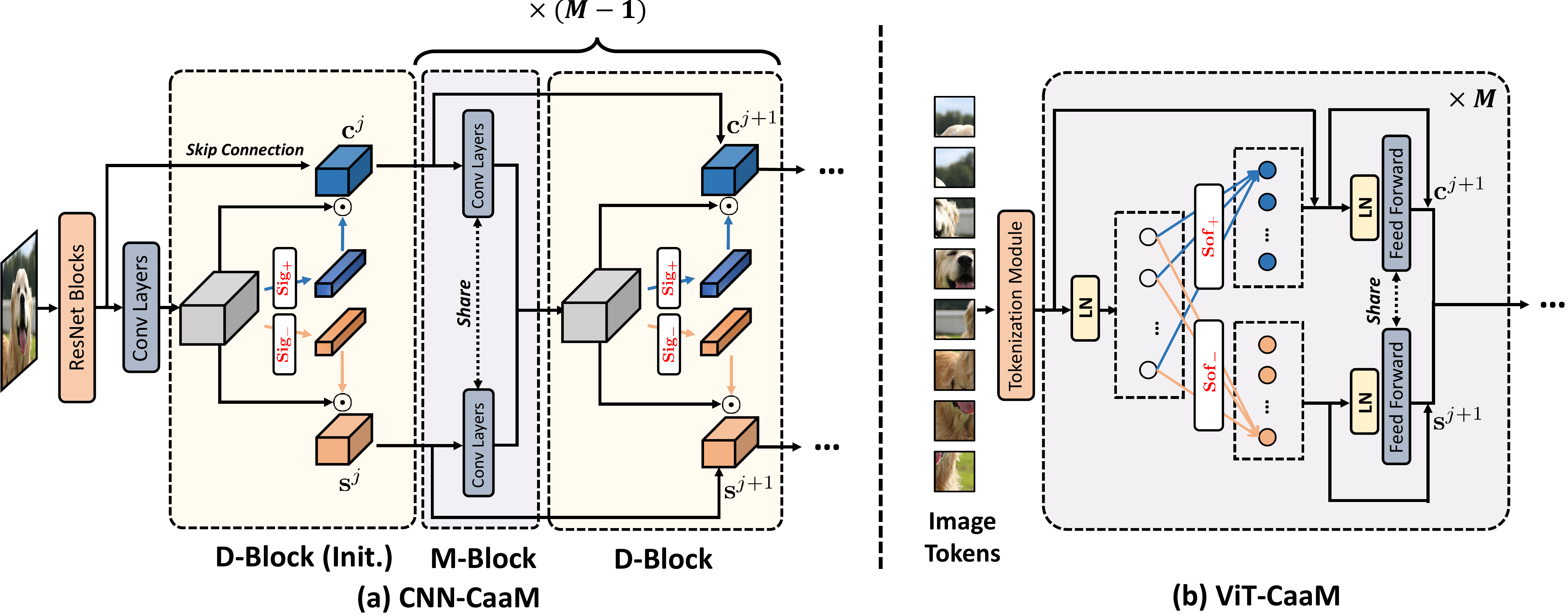}
\end{center}
\vspace{-3mm}
  \caption{
  The network architectures of our CNN-CaaM based on CBAM~\cite{woo2018cbam} and our ViT-CaaM based on T2T-ViT~\cite{yuan2021tokens}.
{\color{red}Red} formulas are used to generate our complementary attentions: {\color{red}$\text{Sig}_{+}$} denotes $\text{Sigmoid}(\mathbf{z})$ and {\color{red}$\text{Sig}_{-}$} for $\text{Sigmoid}(\mathbf{-z})$. {\color{red}$\text{Sof}_{+}$} denotes $\text{Softmax}(\mathbf{q}\mathbf{k}^{\mathrm{T}}/\sqrt{d_K})$ and {\color{red}$\text{Sof}_{-}$} for $\text{Softmax}(-\mathbf{q}\mathbf{k}^{\mathrm{T}}/\sqrt{d_K})$.
For CNN-CaaM, D-Block is utilized to disentangle causal feature $\mathbf{c}$ (blue) and confounder feature $\mathbf{s}$ (orange) from the CNN feature $\mathbf{x}$. D-Block (Init.) denotes the first D-Block.
While M-Block merges $\mathbf{c}$ and $\mathbf{s}$ with the convolution layer.
Then M-Block and D-Block are stacked to progressively refine $\mathbf{c}$ and $\mathbf{s}$.
}
\vspace{-0.4cm}
\label{fig:figure_framework_all}
\end{figure*}

\subsubsection{CNN-CaaM}

CBAM~\cite{woo2018cbam} sequentially adopts the channel and spatial attention module for adaptive CNN feature refinement --- one of the most fundamental ways of computing attention in CNN. 
Given an input feature $\mathbf{x}$,
the attention feature $\mathbf{x'}$ is computed as:
{\small
\begin{equation}
    \mathbf{z}=\text{CBAM}(\mathbf{x}), ~~~
    \mathbf{x'}=\text{sigmoid}(\mathbf{z})\odot \mathbf{x},
    \label{eq:CBAM}
\end{equation}}
where $\mathbf{z} \in \mathbb{R}^{w\times h\times c}$ and $\odot$ denotes the element-wise product.
Therefore, our CaaM attention calculus based on CBAM is defined as:
{\small
\begin{equation}
\textbf{CaaM}:
\begin{cases}
    \mathbf{z}=\text{CBAM}(\mathbf{x}),\\
    \mathbf{c}=\text{Sigmoid}(\mathbf{z})\odot \mathbf{x},\\
    \mathbf{s}=\text{Sigmoid}(-\mathbf{z})\odot \mathbf{x}
\end{cases}\\
\label{eq:cnn-caam}
\end{equation}}
where $\text{Sigmoid}(\mathbf{z})$ and $\text{Sigmoid}(-\mathbf{z})=1-\text{Sigmoid}(\mathbf{z})$ are complementary 
such as to disentangle $\mathbf{c}$ and $\mathbf{s}$ from the input feature $\mathbf{x}$.
Below we elaborate the details of plugging CaaM in residual blocks, as illustrated in Figure~\ref{fig:figure_framework_all} (a).

\noindent\textbf{Disentanglement Block (D-Block).}
D-Block is the block that contains CaaM calculus to generate two attention features $\mathbf{c}$ and $\mathbf{s}$. Note that before D-Block, there can be any number of standard residual blocks~\cite{he2016deep}.
The formulation of $\text{D-Block}^{j+1}$ with residual connection is thus as follows,
{\small
\begin{equation}
    \textbf{D-Block}^{j+1}:
    \begin{cases}
    \mathbf{\hat{c}}^{j}, \mathbf{\hat{s}}^{j} = \text{CaaM}(\mathbf{x}^{j}),\\
    \mathbf{c}^{j+1} = \mathbf{\hat{c}}^{j+1} +  \mathbf{c}^{j}~~~({\footnotesize\text{Skip Connection}}), \\
    \mathbf{s}^{j+1} = \mathbf{\hat{s}}^{j+1} +  \mathbf{s}^{j}~~~({\footnotesize\text{Skip Connection}})    
    \end{cases}\\
    \label{eq:fd_block}
\end{equation}}
where $\mathbf{x}^{j}$ is the feature output by the $j$-th residual block, and
note that as shown in Figure~\ref{fig:figure_framework_all} (a), the first D-Block is denoted as D-Block (Init.), which is slightly different from the following D-Block: 1) The skip connection is connected from the output of the standard ResNet blocks. 2) We remove skip connection on confounder feature $\mathbf{s}^{j}$ to distinguish it from the causal feature $\mathbf{c}^{j}$.

\noindent\textbf{Merge Block (M-Block).}
As shown in Figure~\ref{fig:figure_framework_all}~(a), before D-Block, $\mathbf{c}$ and $\mathbf{s}$ are input into M-Block for feature fusion to get ready for the following D-Block. 
We denote the M-Block (the one left before D-Block) as $\text{M-Block}^{j}$, where $j+1$ is the index of D-Block, and introduce its formulation as follows:
{\small
\begin{equation}
\textbf{M-Block}^{j}:
    \mathbf{x}^{j} = \text{Conv}(\mathbf{c}^{j})+\text{Conv}(\mathbf{s}^{j}).
\label{eq:mr_block}
\end{equation}}
Iterating Eq.~\eqref{eq:fd_block} and Eq.~\eqref{eq:mr_block} yields the multi-layer CaaM (M-Block$\rightarrow$D-Block$\rightarrow$M-Block). 
During inference, we use the final causal feature $\mathbf{c}^{j+M-1}$ for robust prediction.

\subsubsection{ViT-CaaM}
We build ViT-CaaM based on an advanced ViT model called Token-to-Token (T2T)-ViT~\cite{yuan2021tokens} in which the T2T module aims to address the issue of simple tokenization in vanilla ViT~\cite{dosovitskiy2020image}. 
Our CaaM is only plugged in the ViT attention modules of T2T-ViT, thus it is suitable for any ViT-based model, \egs, DeiT~\cite{touvron2020training} and VT~\cite{wu2020visual}.

As shown in Figure~\ref{fig:figure_framework_all}~(b), given the input feature $\mathbf{x}\in \mathbb{R}^{n\times d}$, CaaM first computes the query, key and value vectors $\mathbf{q}, \mathbf{k}, \mathbf{v}\in \mathbb{R}^{n\times d_k}$
using the standard self-attention, where $n$ is the number of image patches, and $d$ and $d_k$ are feature dimensions.
Then, it calculates 
the complementary attentions using Softmax functions.
The overall formulation of CaaM in ViT is thus as follows:
{\small
\begin{equation}
\textbf{CaaM}:
\begin{cases}
    \mathbf{q}, \mathbf{k}, \mathbf{v} = \mathbf{W}_q\mathbf{x}, \mathbf{W}_k\mathbf{x}, \mathbf{W}_v\mathbf{x},\\
    \mathbf{c}=\text{Softmax}(\frac{\mathbf{q}\mathbf{k}^{\mathrm{T}}}{\sqrt{d_K}}) \mathbf{v},\\
    \mathbf{s}=\text{Softmax}(-\frac{\mathbf{q}\mathbf{k}^{\mathrm{T}}}{\sqrt{d_K}}) \mathbf{v}
\end{cases}\\
\label{eq:vit-caam}
\end{equation}}

Different from CNN-CaaM, ViT-CaaM does not have D-Block and M-Block due to the transformer architecture.
Given the input feature $\mathbf{x}^{j}$ 
(\ies, the output feature of the $j$-th transformer module), the ${j}$+1-th module disentangles it to be intermediate features ($\mathbf{\hat{c}}^{j+1}$ and $\mathbf{\hat{s}}^{j+1}$) by applying CaaM attentions, as indicated by the blue and yellow links in Figure~\ref{fig:figure_framework_all} (b).
Then, $\mathbf{\hat{c}}^{j+1}$ and $\mathbf{\hat{s}}^{j+1}$ are fed into an MLP to generate the causal and confounder features, \ies, $\mathbf{c}^{l+1}$ and $\mathbf{s}^{l+1}$.
Note that 1) layer norm (LN) and skip connection are applied in every block, following the standard ViT~\cite{dosovitskiy2020image, yuan2021tokens}; and 2) similar to CNN-CaaM, we omit the first skip connection when generating $\mathbf{\hat{s}}^{j+1}$ in order to avoid the entanglement between $\mathbf{\hat{c}}^{j+1}$ and $\mathbf{\hat{s}}^{j+1}$.
Thus, the basic block of ViT-CaaM can be formulated as follows:
{\small
\begin{equation}
\begin{cases}
    \mathbf{\hat{c}}^{j+1}, \mathbf{\hat{s}}^{j+1} = \text{CaaM}(\text{LN}(\mathbf{x}^{j})),\\
    \mathbf{c}^{j+1} = \text{MLP}(\text{LN}(\mathbf{\hat{c}}^{j+1} + \mathbf{x}^{j}))+\mathbf{\hat{c}}^{j+1} + \mathbf{x}^{j},\\ 
    \mathbf{s}^{j+1} = \text{MLP}(\text{LN}(\mathbf{\hat{s}}^{j+1}))+\mathbf{\hat{s}}^{j+1},\\
    \mathbf{x}^{j+1} = \mathbf{c}^{j+1}+\mathbf{s}^{j+1}
\end{cases} \\
\label{eq:mr_transformer}
\end{equation}}
Iterating Eq.~\eqref{eq:mr_transformer} yields the multi-layer ViT-CaaM. We use the final causal feature $\mathbf{c}^{j+M}$ for prediction in inference.

\section{Experiments}\label{sec:exp}

\subsection{Datasets and Settings}

\noindent\textbf{NICO~\cite{he2021towards}}
\label{sec:nicodata}
is a real-world image dataset designed for OOD settings.
It contains 19 object classes, 188 contexts and nearly 25,000 images in total.
Each image has an object label as well as a context label.
It is thus convenient to ``shift'' the distribution of a class, \ie, by adjusting the proportions of specific contexts for training and testing samples.

\noindent\textbf{Our Settings:} We use the animal subset of NICO. For each animal class, we randomly sample its images and make sure the context labels of those images are within a fixed set of 10 classes (\egs, ``snow'', ``on grass'' and ``in water'').
Based on these data, we propose a challenging OOD setting including three factors regarding contexts: 1)~Long-Tailed---training context labels are in long-tailed distribution in each individual class, \egs, ``sheep'' might have 10 images of ``on grass'', 5 images of ``in water'' and 1 image of ``on road'';
2)~Zero-Shot---for each object class, 7 out of 10 context labels are in training images and the other 3 labels appear only in testing; 
and 3) Orthogonal---the head context label of each object class is set to be as unique (dominating only in one object class) as possible. 
Please kindly refer to the Appendix for more details of our settings.

\noindent\textbf{ImageNet-9~\cite{ilyas2019adversarial}}
Following the related work~\cite{bahng2020learning}, we also evaluate our models on ImageNet-9,
which is a subset of ImageNet containing 9 super-classes with 54,600/2,100 training and validation samples.

\noindent\textbf{Our Settings:} We have three settings to evaluate our model performance on the ImageNet-9.
1) Biased---This is a conventional metric that the accuracy is measured on the whole validation set, serving as a in-distribution testing.
2) Unbiased---This is taken as a proxy to the perfectly debiased test data. To achieve it, we follow~\cite{bahng2020learning} to categorize images into different contexts (\ie, assign context labels to images) by clustering image textures into several groups.
We compute the accuracy for each image cluster and average these accuracies as the final unbiased metric.
3) ImageNet-A~\cite{hendrycks2019natural}. 
It was proposed as a particularly challenging OOD testset of ImageNet. It contains 7,500 real-world images that fool the image classifiers trained on standard ImageNet.
Such ``fool'' was caused by \underline{confounders}---
``the model heavily rely on the \underline{colors}, \underline{textures} and frequently appearing \underline{backgrounds}''~\cite{hendrycks2019natural}.
Therefore, this testing set exactly validates our model performance of deconfounding.
Please kindly refer to the Appendix for more details.

\begin{table*}[t]
\centering
\scalebox{0.73}{
\begin{tabular}{clcccccccccc}
\toprule\toprule
\multicolumn{2}{c}{\multirow{3}{*}{\large{Model}}}       & \multicolumn{5}{c}{CNN-Based}                                                       & \multicolumn{5}{c}{ViT-Based}                                                       \\ \cmidrule(lr){3-7} \cmidrule(lr){8-12}
\multicolumn{2}{c}{}                             & \multicolumn{2}{c}{NICO}        & \multicolumn{2}{c}{ImageNet-9~\cite{ilyas2019adversarial}}         & ImageNet-A~\cite{hendrycks2019natural}        &  
\multicolumn{2}{c}{NICO}        & \multicolumn{2}{c}{ImageNet-9~\cite{ilyas2019adversarial}}     & ImageNet-A~\cite{hendrycks2019natural}              \\ \cmidrule(lr){3-4} \cmidrule(lr){5-6} \cmidrule(lr){7-7} \cmidrule(lr){8-9} \cmidrule(lr){10-11} \cmidrule(lr){12-12}
\multicolumn{2}{c}{}                             & Val            & Test           & Biased         & Unbiased~\cite{bahng2020learning}       & Test           & Val            & Test           & Biased         & Unbiased~\cite{bahng2020learning}       & Test           \\ \midrule
\multirow{5}{*}{\rotatebox{90}{\large{Conv.}}}  & ResNet18~\cite{he2016deep}             & 43.77          & 42.61          & 95.00          & 94.40           & 33.67          & --          & --          & --          & --          & --          \\
& ResNet18+CBAM~\cite{woo2018cbam} &  42.15     &  42.46    & 94.81          & 94.09     & 34.31     & --          & --          & --          & --          & -- \\
& T2T-ViT7~\cite{yuan2021tokens} & --          & --          & --          & --          & --         & 36.23          & 35.62          & 88.76          & 88.35          & 31.28 \\
                        & RUBi~\cite{cadene2019rubi}                   & 43.86          & 44.37          & 94.81          & 94.27          & 34.13          & 35.27          & 34.15          & 87.95               & 87.48               & 29.90               \\
                        & ReBias~\cite{bahng2020learning}                 & 44.92          & 45.23          & 95.20           & 94.89          & 34.26          & 35.28          & 35.74          & 88.99               &    88.32            & 29.33               \\ 
                        & Cutout~\cite{devries2017improved}    & 43.69  & 43.77  & 95.24  & 94.81  & 34.68  & 35.31  & 33.69  & 87.52  & 86.47  & 27.97  \\ 
                        & Mixup~\cite{zhang2017mixup}  & 44.85  & 41.46  & 95.43  & 94.79  & {\color{blue}\textbf{37.71}}  & 37.85  & 34.31  & 89.72  & 88.66  &  30.73  \\ \midrule
\multirow{4}{*}{\rotatebox{90}{\small{w/ H.A. $\mathcal{T}$}}}   & IRM~\cite{arjovsky2019invariant}                    & 40.62          & 41.46          & 94.13          & 94.41          & 33.52          & 36.46          & 34.38          & 89.43          & 88.87          & 30.17          \\
                        & REx~\cite{krueger2020out}                    & 41.00          & 41.15          & 94.15          & 94.28          & 33.18          & 36.23          & 33.46          & 88.52          & 87.26          & 29.18          \\
                        & Unshuffle~\cite{teney2020unshuffling}              & 43.15          & 43.00          & 94.71          & 94.33          & 34.41          & 37.38          & 36.00          & 87.38          & 86.86          & 28.61          \\
                        & \textbf{CaaM (Ours)}     & \cellcolor{mygray}{\color{blue}\textbf{45.46}}          & \cellcolor{mygray}{\color{blue}\textbf{45.77}}          & \cellcolor{mygray}{\color{blue}\textbf{95.52}}          & \cellcolor{mygray}{\color{blue}\textbf{94.96}}          & \cellcolor{mygray}{35.60}          & \cellcolor{mygray}{\color{red}\textbf{38.08}} & \cellcolor{mygray}{\color{blue}\textbf{37.54}}          & \cellcolor{mygray}{\color{blue}\textbf{90.05}}          & \cellcolor{mygray}{\color{blue}\textbf{89.35}}          & \cellcolor{mygray}{\color{blue}\textbf{32.01}} \\ \midrule
\multirow{4}{*}{\rotatebox{90}{\small{w/o H.A. $\mathcal{T}$}}} & IRM~\cite{arjovsky2019invariant}                    & 40.54          & 41.23          & 94.09          & 94.32          & 33.39          & 33.76          & 33.77          & 89.62          & 88.98          & 29.25          \\
                        & REx~\cite{krueger2020out}                    & 40.85          & 41.52          & 93.26          & 93.79          & 32.84          & 35.62          & 34.00          & 88.68          & 87.01          & 28.72          \\
                        & Unshuffle~\cite{teney2020unshuffling}              & 41.69          & 41.61          & 94.81          & 94.30          & 34.04          & 33.62          & 32.92          & 88.38          & 87.39          & 28.52          \\
                        & \textbf{CaaM (Ours)} & \cellcolor{mygray}{\color{red}\textbf{46.38}} & \cellcolor{mygray}{\color{red}\textbf{46.62}} & \cellcolor{mygray}{\color{red}\textbf{96.19}} & \cellcolor{mygray}{\color{red}\textbf{95.83}} & \cellcolor{mygray}{\color{red}\textbf{38.55}} & \cellcolor{mygray}{\color{blue}\textbf{38.00}}          & \cellcolor{mygray}{\color{red}\textbf{37.61}} & \cellcolor{mygray}{\color{red}\textbf{90.33}} & \cellcolor{mygray}{\color{red}\textbf{90.01}} & \cellcolor{mygray}{\color{red}\textbf{32.38}}          \\ \bottomrule\bottomrule
\end{tabular}}
\vspace{1mm}
\caption{Recognition accuracies (\%) based on ResNet18 and T2T-ViT7, on the NICO, ImageNet-9 and ImageNet-A datasets. ``Conv.'', ``w/ H.A. $\mathcal{T}$'', ``w/o H.A. $\mathcal{T}$'' denote conventional methods, causal intervention with human-annotated partitions $\mathcal{T}$ (\ies, ground truth context splits) and intervention without partition annotations, respectively. Our results are highlighted. The {\color{red}\textbf{best}} and {\color{blue}\textbf{second best}} accuracies are marked for all settings.}
\label{tab:sota}
\vspace{-0.4cm}
\end{table*}

\subsection{Implementation Details}

We implement CaaM on two backbones: ResNet18~\cite{he2016deep} for CNN-CaaM and T2T-ViT7~\cite{yuan2021tokens} for ViT-CaaM.
Compared to the original ViT, T2T-ViT introduces a layer-wise Tokens-to-token (T2T) transformation to progressively structurize the image to tokens by aggregating neighboring information.
Below we introduce the comparable baseline methods, \ies, debias methods and intervention methods.

\noindent\textbf{Debias Methods.}
We compare our CaaM with two SOTA methods: RUBi~\cite{cadene2019rubi} and ReBias~\cite{bahng2020learning}.
RUBi explicitly learns a biased model using biased input, and then performs de-biasing on a standard model by re-weighting its prediction logits where weights are generated by the biased model. 
The other SOTA is ReBias. It utilizes a small receptive fields CNN (BagNet~\cite{brendel2019approximating}) to explicitly encode context bias in a model, and the debiased representation is encouraged to be statistically independent from it.

\noindent\textbf{Intervention Methods.}
We compare our CaaM with three SOTA causal intervention methods: IRM~\cite{arjovsky2019invariant}, REx~\cite{krueger2020out} and Unshuffle~\cite{teney2020unshuffling}.
IRM claimed that the robust representation derives the image classifiers who can make invariant predictions across different contexts (for the same object class).
REx is an improved version of IRM and its key is to encourage robustness over affine combinations for the training risks.
Unshuffle~\cite{teney2020unshuffling} extended IRM to a real vision-language tasks---visual question answering.
It trains an individual classifier for each data split and applies a variance regularizer on classifier weights.

Note that all above methods require the annotation of data splits: either from manual labeling ~\cite{arjovsky2019invariant, chang2020invariant,ahuja2020invariant} or from pre-defined clustering~\cite{teney2020unshuffling, bahng2020learning}.
While our CaaM does not need such annotations, leading to the intervention in an unsupervised fashion.
To further show our superiority, we conduct two groups of comparisons to intervention methods (in the bottom blocks of Table~\ref{tab:sota}): one allows all models to be trained with human annotated splits (w/ H.A. $\mathcal{T}$), and the other one without (w/o H.A. $\mathcal{T}$).
On the NICO dataset, H.A. $\mathcal{T}$ is 
built by using context labels, \ies, images containing same context are in the same split.
On the ImageNet-9 dataset, H.A. $\mathcal{T}$ is built by clustering context features into several splits, following~\cite{bahng2020learning}.

\subsection{Comparing to SOTAs}

Table~\ref{tab:sota} shows all comparisons we conduct on the NICO, ImageNet-9 and ImageNet-A(test only) datasets. 
It is obvious that our CaaM achieves the top performance across all settings. 
It is worth highlighting that 1) in the setting ``w/ H.A. $\mathcal{T}$'', 
our CaaM surpasses intervention methods by clear margins, \egs, $2.1\%$ and $1.2\%$ higher than IRM and Unshuffle, respectively, on the challenging ImageNet-A (with CNN); and 2) these margins are even larger, \egs, increased to $5\%$ for ImageNet-A (with CNN), in the more difficult setting ``w/o H.A. $\mathcal{T}$'', where ours get improved but others' are reduced.
These validate that our self-annotating for partitions---progressively updating $\mathcal{T}$ using CaaM---indeed achieves the superior representations of confounders than using any hard or manual splits as in~\cite{arjovsky2019invariant, teney2020unshuffling,krueger2020out}.

\begin{table}[]
\centering
\scalebox{0.8}{
\begin{tabular}{llcccc}
\toprule\toprule
\multicolumn{2}{c}{\multirow{2}{*}{Settings}}           & \multicolumn{2}{c}{CNN-CaaM}                          & \multicolumn{2}{c}{ViT-CaaM}                          \\ \cline{3-6} 
\multicolumn{2}{c}{}                                    & Val                       & Test                      & Val                       & Test                      \\ \hline
\multirow{3}{*}{{\rotatebox{90}{\small{\textbf{Num L.}}}}}     & $M$=1              & 43.92                     & 44.54                     & 35.54                     & 36.77                     \\
                                   & $M$=2              & 45.46                     & 45.77                     & 37.89                     & 37.46                     \\
                                   & $M$=4              & 44.15                     & 45.31                     & \textbf{38.08}            & 37.54                     \\ \hline
\multirow{3}{*}{{\rotatebox{90}{\small{\textbf{Num S.}}}}}     & $m$=2              & \multicolumn{1}{l}{43.98} & \multicolumn{1}{l}{44.92} & 37.87                     & 37.32                     \\
                                   & $m$=4              & 45.46                     & 45.77                     & \textbf{38.08}            & 37.54                     \\
                                   & $m$=8              & \multicolumn{1}{l}{45.23} & \multicolumn{1}{l}{45.74} & \multicolumn{1}{l}{37.94} & \multicolumn{1}{l}{37.23} \\ \hline
\multirow{3}{*}{{\rotatebox{90}{\small{\textbf{T.S.}}}}} & Reboot Training    & \multicolumn{1}{l}{44.23} & \multicolumn{1}{l}{44.46} & \multicolumn{1}{l}{36.69} & \multicolumn{1}{l}{35.46} \\
                                   & Randomize $\theta$ & 44.46                     & 43.38                     & 37.12                     & 36.15                     \\
                                   & CaaM               & \textbf{46.38}            & \textbf{46.62}            & 38.00                     & \textbf{37.61}            \\ \bottomrule\bottomrule
\end{tabular}}
\vspace{2mm}
\caption{Ablation studies of our proposed CNN-CaaM and ViT-CaaM on NICO dataset. \textbf{Num L.}, \textbf{Num S.} and \textbf{T.S.} denote number of layers, number of splits and training schedule respectively.}
\label{tab:ablation1}
\vspace{-0.4cm}
\end{table}

\subsection{Ablation Study}
\label{sec:ablation_study}

\noindent\textbf{Q1:} \textit{What are the optimal hyperparameters of CaaM?}
We replace the last $M$ blocks of CNN (or the last $M$ layers of ViT) and vary of value of $M$ to find out how many attention blocks we need to get the best performance. 
Similarly, we use $m$ context splits and vary its values to find out the optimal number of partitions. 

\noindent\textbf{A1:} We can see from the top block of Table~\ref{tab:ablation1} that the performance of CaaM saturates around $M$=2 for CNN-CaaM ($M$=4 for ViT-CaaM). As we add CaaM layers from top to bottom, this is perhaps because the lower CNN feature maps do not emerge foreground and background semantics yet.
On the middle block of Table~\ref{tab:ablation1}, we find that $m$=4 is the best but the accuracy differences with other values are not significant, \ies, not sensitive to $m$. 

\noindent\textbf{Q2:} \textit{What is the advantage of progressively updating and aggregating the effects of different partitions? Is the Maxi-Game indispensable?}
We conduct two ablation studies: one is to omit the optimization in Eq.~\eqref{eq:jointloss} and each phase we randomize the weights of those parameters, \egs, $\mathcal{A}$ and $\bar{\mathcal{A}}$, which we denote as ``Reboot Training''; the other is to omit the optimization in Eq.~\eqref{eq:maxigame}, and similarly, each phase we randomize the weights of $\theta$, \ies, ``Randomize $\theta$''.

\begin{figure*}[t]
\begin{center}
\includegraphics[width=0.93\textwidth]{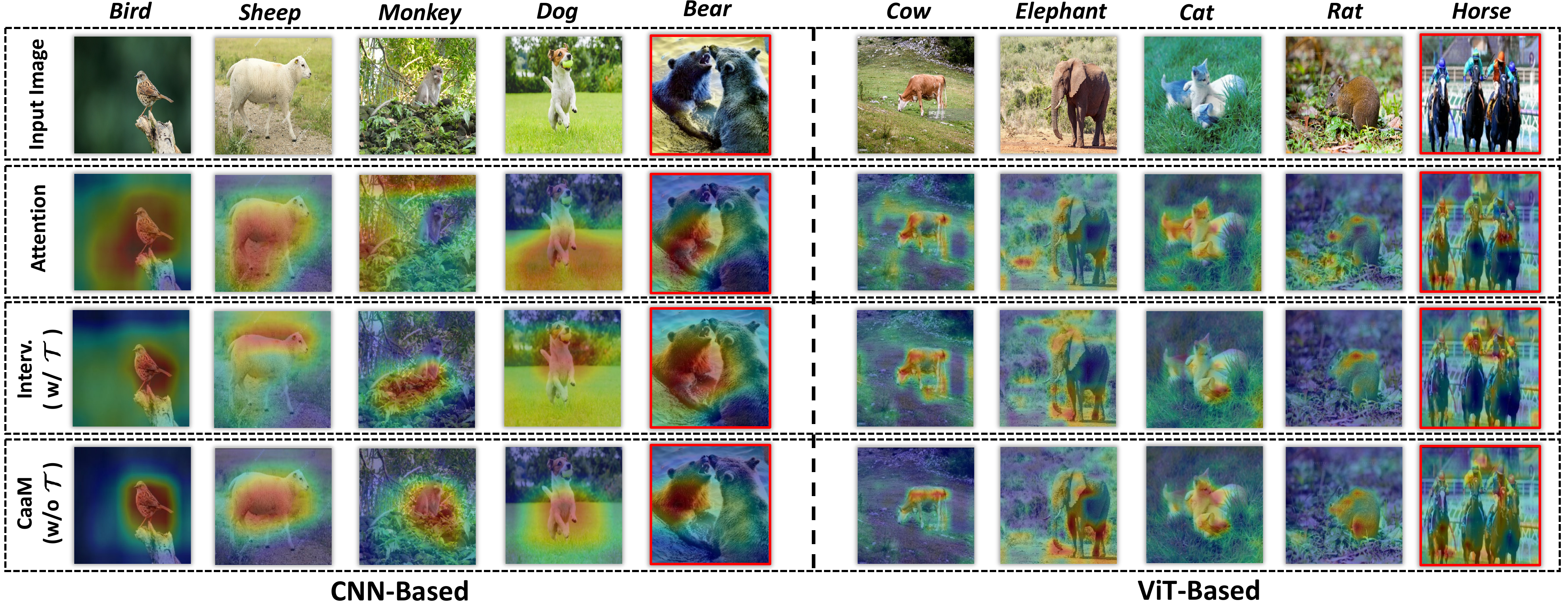}
\vspace{-0.4cm}
\end{center}
  \caption{Visualization of the attention map with our CaaM and baseline methods based on CNN and ViT. ``Attention'' and ``Interv.'' denote the conventional attention model and intervention method~\cite{teney2020unshuffling} respectively. The red box represents the failure case.}
\vspace{-0.4cm}
\label{fig:quali2}
\end{figure*}

\begin{figure}[t]
\begin{center}
\includegraphics[width=0.48\textwidth]{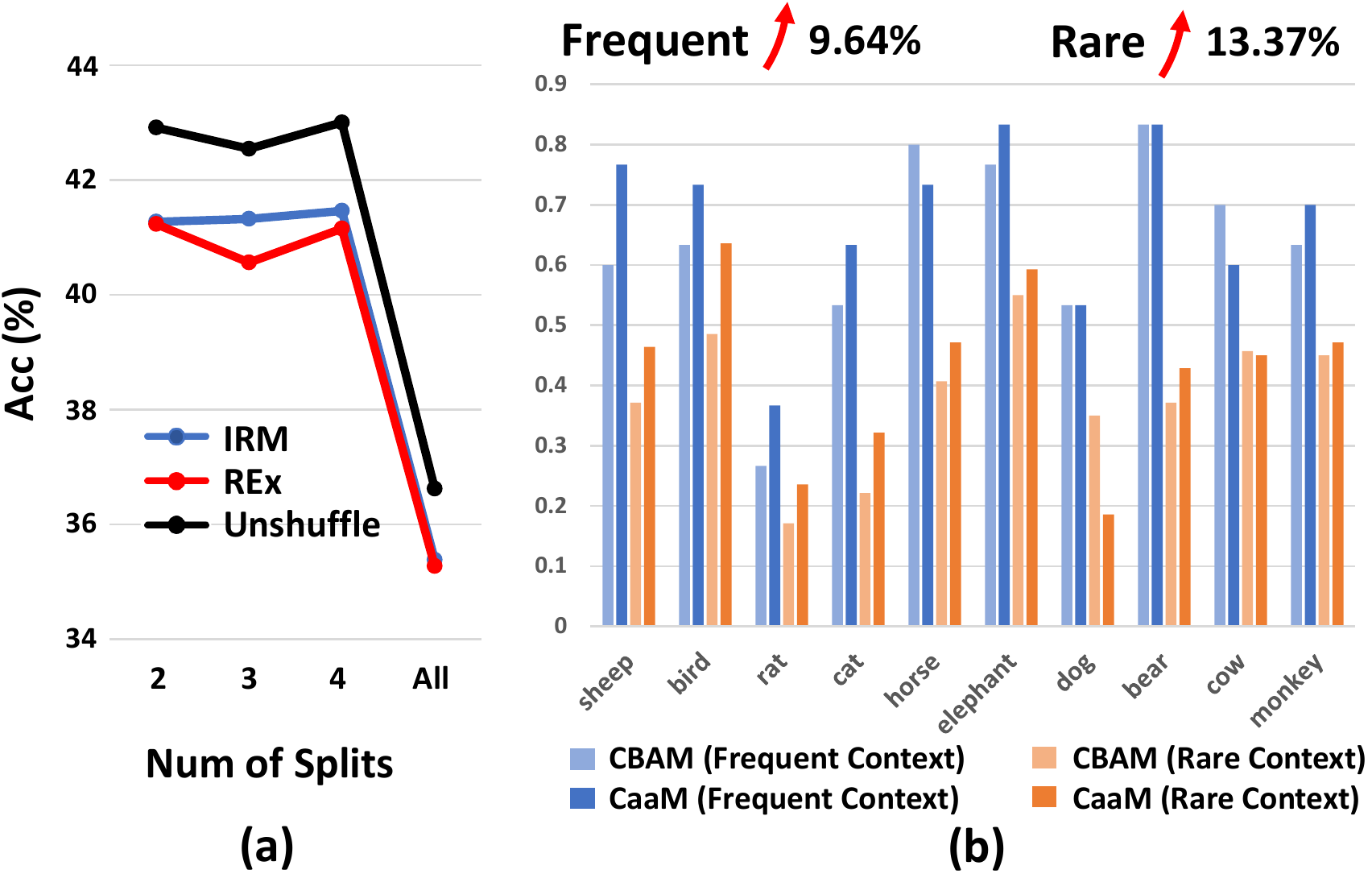}
\end{center}
\vspace{-0.4cm}
  \caption{(a) The performance of intervention methods with different number of splits on NICO dataset. ``All'' denotes grouping data according to each context, \ies, \#Num of Splits$=$\#Num of Contexts. (b) The per-class classification accuracy comparison between conventional CBAM~\cite{woo2018cbam} and our CaaM on images with frequent and rare context respectively 
  .}
\vspace{-0.4cm}
\label{fig:quali}
\end{figure}

\noindent\textbf{A2:}
We show the corresponding results on the bottom block of Table~\ref{tab:ablation1}. 
It is clear that ``Reboot Training'' in each phase results in sharp performance drops for all models, by $2\%$ on average.
Besides, ``Randomize $\theta$'' also brings clear performance drops while the reduced margins are smaller than those of ``Reboot Training''.
These validate that our CaaM is a organic integrity composed of collaborative and optimizable modules.

\begin{table}[t]
\centering
\scalebox{0.8}{
\begin{tabular}{clcc}
\toprule\toprule
Setting                   & \multicolumn{1}{c}{Model} & CNN-Based & ViT-Based \\ \hline
                          & Attention                 & 69.73     & 55.36     \\ \midrule
\multirow{2}{*}{w/ Partition} & Interv.  ~\cite{teney2020unshuffling}            & 73.61     &  56.71     \\
                          & \textbf{CaaM (Ours)}               & 77.52     & 58.24     \\ \midrule
w/o Partition                 & \textbf{CaaM (Ours)}               & \textbf{78.37}     & \textbf{58.83}      \\ \bottomrule\bottomrule
\end{tabular}}
\vspace{2mm}
\caption{The attention map accuracy (\%) of using different models on ImageNet-9 test set with ground truth bounding boxes. ``Attention'' denotes the conventional attention model, \ies, ResNet+CBAM and ViT. ``Interv.'' is the abbreviation of the intervention method~\cite{teney2020unshuffling}.}
\label{table:attn_acc}
\vspace{-0.4cm}
\end{table}

\noindent\textbf{Q3:} \textit{Can CaaM achieve robust attention?}
To evaluate the exactness of the attention map generated by CaaM quantitatively, we calculate the attention accuracy of our CaaM and baseline methods with the groud truth object bounding box coordinates of ImageNet-9 test set.
Specifically, the attention accuracy is given by the ratio between the attention area in bounding box and the whole attention area. Details are given in Appendix.

\noindent\textbf{A3:}
We report the attention accuracy in Table~\ref{table:attn_acc}. Compared to the conventional attention, the intervention method with ground truth context partition can achieve better performance; while our CNN-CaaM and ViT-CaaM largely outperform these two methods in both settings (\ies, with and without partitions). This result fully demonstrates the both effectiveness of our multi-layer complementary attention and the adversarial training pipeline.

\noindent\textbf{Q4:} \textit{Why merge the ground truth contexts into bigger splits?}
Recall that in Section~\ref{sec:intro}, we explain that the context absence of a class violates the positivity assumption.
To evaluate the effect, we provide the detailed results on the NICO dataset with different number of data splits in Figure~\ref{fig:quali} (a).

\noindent\textbf{A4:}
We can see that the accuracies of the intervention keep relatively stable for $2-4$ splits, but have a huge drop when grouping according to each context due to the violation.

\noindent\textbf{Q5:} \textit{Does CaaM boost the recognition of both the samples with frequent contexts and with rare contexts?} We show the performance of our CaaM and conventional CBAM attention model on test samples with frequent context and rare context respectively in Figure~\ref{fig:quali} (b).

\noindent\textbf{A5:}
Specifically, the ``frequent'' denotes the top three context classes in training distribution, while ``rare'' represents the tailed seven context classes containing three zero-shot classes.
Recall that as shown in Figure~\ref{fig:figure1}, the performance decreases more for the rare context classes using conventional attention model. Conversely, our CaaM can even receive a greater performance boost on rare contexts (13.37\%) than that of the frequent (9.64\%).
Moreover, in relevant research fields (\egs, long-tailed classification), it is well-known that the improvement of tail (rare) classes usually sacrifices the performance of the head (frequent). However, our CaaM can improve the accuracy of frequent and rare context simultaneously with a large margin.

\subsection{Qualitative Results.}

Figure~\ref{fig:quali2} shows the qualitative attention map comparisons between our proposed CaaM (without the partition $\mathcal{T}$), intervention methods (with known partition $\mathcal{T}$) and convention attention model, \ies, CBAM~\cite{woo2018cbam} and T2T-ViT7~\cite{yuan2021tokens}.
Note that current ViT models are limited that the attention mechanism cannot be well trained without large-scale dataset. 
For attention visualization, the weights of T2T-ViT7 are initialized with ImageNet pretrained models.
From Figure~\ref{fig:quali2} we can see that, compared to the conventional attention (second row) and intervention methods (third row), our CaaM can achieve more accurate attention activation.
Red boxes denote the failure cases. We find that our CaaM also cannot accurately attend to multiple objects (\egs, two bears) or the single object co-existing with other ones (\egs, horses and people). This inspires us to perform surrounding objects adjustment~\cite{wang2020visual, zhang2020causal} in future.
\section{Conclusion}
We demonstrated that the conventional attention module is particularly biased in OOD settings. We postulated that the reason is due to the confounder, whose effect should be removed by causal intervention. We theoretically showed that existing context-invariant methods suffer from improper causal intervention, which can be addressed by the proposed CaaM. Extensive experiments on three challenging benchmarks empirically demonstrated the effectiveness of CaaM. In future, we will seek a more powerful theory of causal effect disentanglement~\cite{higgins2018towards} and its implementations.

\noindent\textbf{Acknowledgement.}
The authors would like to thank all reviewers for their constructive suggestions. This research is partly supported by the Alibaba-NTU Joint Research Institute, the A*STAR under its AME YIRG Grant (Project No. A20E6c0101), and the Singapore Ministry of
Education (MOE) Academic Research Fund (AcRF) Tier 2
grant.

{\small
\bibliographystyle{ieee_fullname}
\bibliography{egbib}

\begin{thebibliography}{10}\itemsep=-1pt

\bibitem{achille2018information}
Alessandro Achille and Stefano Soatto.
\newblock Information dropout: Learning optimal representations through noisy
  computation.
\newblock {\em IEEE TPAMI}, 40(12):2897--2905, 2018.

\bibitem{ahuja2020invariant}
Kartik Ahuja, Karthikeyan Shanmugam, Kush Varshney, and Amit Dhurandhar.
\newblock Invariant risk minimization games.
\newblock In {\em ICML}, pages 145--155. PMLR, 2020.

\bibitem{ahuja2020empirical}
Kartik Ahuja, Jun Wang, Amit Dhurandhar, Karthikeyan Shanmugam, and Kush~R
  Varshney.
\newblock Empirical or invariant risk minimization? a sample complexity
  perspective.
\newblock {\em arXiv preprint}, 2020.

\bibitem{arjovsky2019invariant}
Martin Arjovsky, L{\'e}on Bottou, Ishaan Gulrajani, and David Lopez-Paz.
\newblock Invariant risk minimization.
\newblock {\em arXiv preprint}, 2019.

\bibitem{bahng2020learning}
Hyojin Bahng, Sanghyuk Chun, Sangdoo Yun, Jaegul Choo, and Seong~Joon Oh.
\newblock Learning de-biased representations with biased representations.
\newblock In {\em ICML}, pages 528--539. PMLR, 2020.

\bibitem{ben2007analysis}
Shai Ben-David, John Blitzer, Koby Crammer, Fernando Pereira, et~al.
\newblock Analysis of representations for domain adaptation.
\newblock {\em NeurIPS}, 19:137, 2007.

\bibitem{brendel2019approximating}
Wieland Brendel and Matthias Bethge.
\newblock Approximating cnns with bag-of-local-features models works
  surprisingly well on imagenet.
\newblock {\em ICLR}, 2019.

\bibitem{cadene2019rubi}
Remi Cadene, Corentin Dancette, Hedi Ben-Younes, Matthieu Cord, and Devi
  Parikh.
\newblock Rubi: Reducing unimodal biases in visual question answering.
\newblock {\em NeurIPS}, 2019.

\bibitem{chang2020invariant}
Shiyu Chang, Yang Zhang, Mo Yu, and Tommi Jaakkola.
\newblock Invariant rationalization.
\newblock In {\em ICML}, pages 1448--1458. PMLR, 2020.

\bibitem{chen2017sca}
Long Chen, Hanwang Zhang, Jun Xiao, Liqiang Nie, Jian Shao, Wei Liu, and
  Tat-Seng Chua.
\newblock Sca-cnn: Spatial and channel-wise attention in convolutional networks
  for image captioning.
\newblock In {\em CVPR}, pages 5659--5667, 2017.

\bibitem{clark2019don}
Christopher Clark, Mark Yatskar, and Luke Zettlemoyer.
\newblock Don't take the easy way out: Ensemble based methods for avoiding
  known dataset biases.
\newblock {\em EMNLP}, 2019.

\bibitem{corbetta2002control}
Maurizio Corbetta and Gordon~L Shulman.
\newblock Control of goal-directed and stimulus-driven attention in the brain.
\newblock {\em Nat. Rev. Neurosci.}, 3(3):201--215, 2002.

\bibitem{creager2020environment}
Elliot Creager, J{\"o}rn-Henrik Jacobsen, and Richard Zemel.
\newblock Environment inference for invariant learning.
\newblock In {\em ICML Workshop on Uncertainty and Robustness}, 2020.

\bibitem{devlin2018bert}
Jacob Devlin, Ming-Wei Chang, Kenton Lee, and Kristina Toutanova.
\newblock Bert: Pre-training of deep bidirectional transformers for language
  understanding.
\newblock {\em arXiv preprint}, 2018.

\bibitem{devries2017improved}
Terrance DeVries and Graham~W Taylor.
\newblock Improved regularization of convolutional neural networks with cutout.
\newblock {\em arXiv preprint arXiv:1708.04552}, 2017.

\bibitem{dosovitskiy2020image}
Alexey Dosovitskiy, Lucas Beyer, Alexander Kolesnikov, Dirk Weissenborn,
  Xiaohua Zhai, Thomas Unterthiner, Mostafa Dehghani, Matthias Minderer, Georg
  Heigold, Sylvain Gelly, et~al.
\newblock An image is worth 16x16 words: Transformers for image recognition at
  scale.
\newblock {\em arXiv preprint}, 2020.

\bibitem{ganin2016domain}
Yaroslav Ganin, Evgeniya Ustinova, Hana Ajakan, Pascal Germain, Hugo
  Larochelle, Fran{\c{c}}ois Laviolette, Mario Marchand, and Victor Lempitsky.
\newblock Domain-adversarial training of neural networks.
\newblock {\em J. Mach. Learn. Res.}, 17(1):2096--2030, 2016.

\bibitem{geirhos2018imagenet}
Robert Geirhos, Patricia Rubisch, Claudio Michaelis, Matthias Bethge, Felix~A
  Wichmann, and Wieland Brendel.
\newblock Imagenet-trained cnns are biased towards texture; increasing shape
  bias improves accuracy and robustness.
\newblock {\em ICLR}, 2018.

\bibitem{gong2016domain}
Mingming Gong, Kun Zhang, Tongliang Liu, Dacheng Tao, Clark Glymour, and
  Bernhard Sch{\"o}lkopf.
\newblock Domain adaptation with conditional transferable components.
\newblock In {\em ICML}, pages 2839--2848. PMLR, 2016.

\bibitem{he2016deep}
Kaiming He, Xiangyu Zhang, Shaoqing Ren, and Jian Sun.
\newblock Deep residual learning for image recognition.
\newblock In {\em CVPR}, pages 770--778, 2016.

\bibitem{he2021towards}
Yue He, Zheyan Shen, and Peng Cui.
\newblock Towards non-iid image classification: A dataset and baselines.
\newblock {\em Pattern Recognit.}, 110:107383, 2021.

\bibitem{hendrycks2016baseline}
Dan Hendrycks and Kevin Gimpel.
\newblock A baseline for detecting misclassified and out-of-distribution
  examples in neural networks.
\newblock {\em ICLR}, 2017.

\bibitem{hendrycks2019natural}
Dan Hendrycks, Kevin Zhao, Steven Basart, Jacob Steinhardt, and Dawn Song.
\newblock Natural adversarial examples.
\newblock {\em CVPR}, 2021.

\bibitem{heo2021rethinking}
Byeongho Heo, Sangdoo Yun, Dongyoon Han, Sanghyuk Chun, Junsuk Choe, and
  Seong~Joon Oh.
\newblock Rethinking spatial dimensions of vision transformers.
\newblock {\em ICCV}, 2021.

\bibitem{hernan2010causal}
Miguel~A Hern{\'a}n and James~M Robins.
\newblock Causal inference, 2010.

\bibitem{higgins2018towards}
Irina Higgins, David Amos, David Pfau, Sebastien Racaniere, Loic Matthey,
  Danilo Rezende, and Alexander Lerchner.
\newblock Towards a definition of disentangled representations.
\newblock {\em arXiv preprint}, 2018.

\bibitem{holland1986statistics}
Paul~W Holland.
\newblock Statistics and causal inference.
\newblock {\em J. Am. Stat. Assoc.}, 81(396):945--960, 1986.

\bibitem{hu2018squeeze}
Jie Hu, Li Shen, and Gang Sun.
\newblock Squeeze-and-excitation networks.
\newblock In {\em CVPR}, pages 7132--7141, 2018.

\bibitem{hu2021distilling}
Xinting Hu, Kaihua Tang, Chunyan Miao, Xian-Sheng Hua, and Hanwang Zhang.
\newblock Distilling causal effect of data in class-incremental learning.
\newblock In {\em CVPR}, pages 3957--3966, 2021.

\bibitem{huang2017arbitrary}
Xun Huang and Serge Belongie.
\newblock Arbitrary style transfer in real-time with adaptive instance
  normalization.
\newblock In {\em ICCV}, pages 1501--1510, 2017.

\bibitem{ilyas2019adversarial}
Andrew Ilyas, Shibani Santurkar, Dimitris Tsipras, Logan Engstrom, Brandon
  Tran, and Aleksander Madry.
\newblock Adversarial examples are not bugs, they are features.
\newblock {\em NeurIPS}, 2019.

\bibitem{khan2017cost}
Salman~H Khan, Munawar Hayat, Mohammed Bennamoun, Ferdous~A Sohel, and Roberto
  Togneri.
\newblock Cost-sensitive learning of deep feature representations from
  imbalanced data.
\newblock {\em EEE Trans. Neural Netw. Learn. Syst.}, 29(8):3573--3587, 2017.

\bibitem{kim2019learning}
Byungju Kim, Hyunwoo Kim, Kyungsu Kim, Sungjin Kim, and Junmo Kim.
\newblock Learning not to learn: Training deep neural networks with biased
  data.
\newblock In {\em CVPR}, pages 9012--9020, 2019.

\bibitem{krueger2020out}
David Krueger, Ethan Caballero, Joern-Henrik Jacobsen, Amy Zhang, Jonathan
  Binas, Dinghuai Zhang, Remi~Le Priol, and Aaron Courville.
\newblock Out-of-distribution generalization via risk extrapolation (rex).
\newblock {\em arXiv preprint}, 2020.

\bibitem{li2019repair}
Yi Li and Nuno Vasconcelos.
\newblock Repair: Removing representation bias by dataset resampling.
\newblock In {\em CVPR}, pages 9572--9581, 2019.

\bibitem{liang2017enhancing}
Shiyu Liang, Yixuan Li, and Rayadurgam Srikant.
\newblock Enhancing the reliability of out-of-distribution image detection in
  neural networks.
\newblock {\em ICLR}, 2018.

\bibitem{liu2021heterogeneous}
Jiashuo Liu, Zheyuan Hu, Peng Cui, Bo Li, and Zheyan Shen.
\newblock Heterogeneous risk minimization.
\newblock {\em arXiv preprint arXiv:2105.03818}, 2021.

\bibitem{mahajan2018exploring}
Dhruv Mahajan, Ross Girshick, Vignesh Ramanathan, Kaiming He, Manohar Paluri,
  Yixuan Li, Ashwin Bharambe, and Laurens Van Der~Maaten.
\newblock Exploring the limits of weakly supervised pretraining.
\newblock In {\em ECCV}, pages 181--196, 2018.

\bibitem{mnih2014recurrent}
Volodymyr Mnih, Nicolas Heess, Alex Graves, and Koray Kavukcuoglu.
\newblock Recurrent models of visual attention.
\newblock {\em NeurIPS}, 2014.

\bibitem{muandet2013domain}
Krikamol Muandet, David Balduzzi, and Bernhard Sch{\"o}lkopf.
\newblock Domain generalization via invariant feature representation.
\newblock In {\em ICML}, pages 10--18. PMLR, 2013.

\bibitem{niu2020counterfactual}
Yulei Niu, Kaihua Tang, Hanwang Zhang, Zhiwu Lu, Xian-Sheng Hua, and Ji-Rong
  Wen.
\newblock Counterfactual vqa: A cause-effect look at language bias.
\newblock {\em CVPR}, 2021.

\bibitem{pearl2009causality}
Judea Pearl.
\newblock {\em Causality}.
\newblock Cambridge university press, 2009.

\bibitem{pearl2016causal}
Judea Pearl, Madelyn Glymour, and Nicholas~P Jewell.
\newblock {\em Causal inference in statistics: A primer}.
\newblock John Wiley \& Sons, 2016.

\bibitem{ramesh2021zero}
Aditya Ramesh, Mikhail Pavlov, Gabriel Goh, Scott Gray, Chelsea Voss, Alec
  Radford, Mark Chen, and Ilya Sutskever.
\newblock Zero-shot text-to-image generation.
\newblock {\em arXiv preprint}, 2021.

\bibitem{rosenfeld2020risks}
Elan Rosenfeld, Pradeep Ravikumar, and Andrej Risteski.
\newblock The risks of invariant risk minimization.
\newblock {\em arXiv preprint arXiv:2010.05761}, 2020.

\bibitem{shen2016relay}
Li Shen, Zhouchen Lin, and Qingming Huang.
\newblock Relay backpropagation for effective learning of deep convolutional
  neural networks.
\newblock In {\em ECCV}, pages 467--482. Springer, 2016.

\bibitem{tang2020long}
Kaihua Tang, Jianqiang Huang, and Hanwang Zhang.
\newblock Long-tailed classification by keeping the good and removing the bad
  momentum causal effect.
\newblock {\em NeurIPS}, 2020.

\bibitem{tang2020unbiased}
Kaihua Tang, Yulei Niu, Jianqiang Huang, Jiaxin Shi, and Hanwang Zhang.
\newblock Unbiased scene graph generation from biased training.
\newblock In {\em CVPR}, pages 3716--3725, 2020.

\bibitem{teney2020unshuffling}
Damien Teney, Ehsan Abbasnejad, and Anton van~den Hengel.
\newblock Unshuffling data for improved generalization.
\newblock {\em arXiv preprint}, 2020.

\bibitem{touvron2020training}
Hugo Touvron, Matthieu Cord, Matthijs Douze, Francisco Massa, Alexandre
  Sablayrolles, and Herv{\'e} J{\'e}gou.
\newblock Training data-efficient image transformers \& distillation through
  attention.
\newblock {\em arXiv preprint}, 2020.

\bibitem{tzeng2017adversarial}
Eric Tzeng, Judy Hoffman, Kate Saenko, and Trevor Darrell.
\newblock Adversarial discriminative domain adaptation.
\newblock In {\em CVPR}, pages 7167--7176, 2017.

\bibitem{vaswani2021scaling}
Ashish Vaswani, Prajit Ramachandran, Aravind Srinivas, Niki Parmar, Blake
  Hechtman, and Jonathon Shlens.
\newblock Scaling local self-attention for parameter efficient visual
  backbones.
\newblock In {\em CVPR}, pages 12894--12904, 2021.

\bibitem{vaswani2017attention}
Ashish Vaswani, Noam Shazeer, Niki Parmar, Jakob Uszkoreit, Llion Jones,
  Aidan~N Gomez, Lukasz Kaiser, and Illia Polosukhin.
\newblock Attention is all you need.
\newblock {\em arXiv preprint}, 2017.

\bibitem{wang2019learning}
Haohan Wang, Zexue He, Zachary~C Lipton, and Eric~P Xing.
\newblock Learning robust representations by projecting superficial statistics
  out.
\newblock {\em ICLR}, 2019.

\bibitem{wang2020visual}
Tan Wang, Jianqiang Huang, Hanwang Zhang, and Qianru Sun.
\newblock Visual commonsense r-cnn.
\newblock In {\em CVPR}, pages 10760--10770, 2020.

\bibitem{wang2018non}
Xiaolong Wang, Ross Girshick, Abhinav Gupta, and Kaiming He.
\newblock Non-local neural networks.
\newblock In {\em CVPR}, pages 7794--7803, 2018.

\bibitem{woo2018cbam}
Sanghyun Woo, Jongchan Park, Joon-Young Lee, and In~So Kweon.
\newblock Cbam: Convolutional block attention module.
\newblock In {\em ECCV}, pages 3--19, 2018.

\bibitem{wu2020visual}
Bichen Wu, Chenfeng Xu, Xiaoliang Dai, Alvin Wan, Peizhao Zhang, Masayoshi
  Tomizuka, Kurt Keutzer, and Peter Vajda.
\newblock Visual transformers: Token-based image representation and processing
  for computer vision.
\newblock {\em arXiv preprint}, 2020.

\bibitem{xu2015show}
Kelvin Xu, Jimmy Ba, Ryan Kiros, Kyunghyun Cho, Aaron Courville, Ruslan
  Salakhudinov, Rich Zemel, and Yoshua Bengio.
\newblock Show, attend and tell: Neural image caption generation with visual
  attention.
\newblock In {\em ICML}, pages 2048--2057. PMLR, 2015.

\bibitem{yang2020deconfounded}
Xu Yang, Hanwang Zhang, and Jianfei Cai.
\newblock Deconfounded image captioning: A causal retrospect.
\newblock {\em arXiv preprint}, 2020.

\bibitem{yang2021causal}
Xu Yang, Hanwang Zhang, Guojun Qi, and Jianfei Cai.
\newblock Causal attention for vision-language tasks.
\newblock {\em arXiv preprint}, 2021.

\bibitem{ye2021adversarial}
Nanyang Ye, Jingxuan Tang, Huayu Deng, Xiao-Yun Zhou, Qianxiao Li, Zhenguo Li,
  Guang-Zhong Yang, and Zhanxing Zhu.
\newblock Adversarial invariant learning.
\newblock In {\em CVPR}, pages 12446--12454, 2021.

\bibitem{yuan2021tokens}
Li Yuan, Yunpeng Chen, Tao Wang, Weihao Yu, Yujun Shi, Francis~EH Tay, Jiashi
  Feng, and Shuicheng Yan.
\newblock Tokens-to-token vit: Training vision transformers from scratch on
  imagenet.
\newblock {\em CVPR}, 2021.

\bibitem{yue2021counterfactual}
Zhongqi Yue, Tan Wang, Qianru Sun, Xian-Sheng Hua, and Hanwang Zhang.
\newblock Counterfactual zero-shot and open-set visual recognition.
\newblock In {\em CVPR}, pages 15404--15414, 2021.

\bibitem{yue2020interventional}
Zhongqi Yue, Hanwang Zhang, Qianru Sun, and Xian-Sheng Hua.
\newblock Interventional few-shot learning.
\newblock {\em NeurIPS}, 2020.

\bibitem{zhang2020causal}
Dong Zhang, Hanwang Zhang, Jinhui Tang, Xiansheng Hua, and Qianru Sun.
\newblock Causal intervention for weakly-supervised semantic segmentation.
\newblock {\em NeurIPS}, 2020.

\bibitem{zhang2017mixup}
Hongyi Zhang, Moustapha Cisse, Yann~N Dauphin, and David Lopez-Paz.
\newblock mixup: Beyond empirical risk minimization.
\newblock {\em arXiv preprint arXiv:1710.09412}, 2017.

\end{thebibliography}
}

\newpage



\section*{Supplementary material}

# Appendix

This appendix is organized as follows:

- Section A.1 provides further interpretations of our proposed CaaM.

- Section A.2 presents theoretical evidences for the **Improper Causal Intervention** (Section 3.1), and for the convergence of **Adversarial Training** (Section 3.2).

- Section A.3 provides additional implementation details for **Invariant Loss** and **Adversarial Training** (Section 3.2).

- Section A.4 provides the methods used to generate OOD datasets (Section 4.1), additional training details (Section 4.2), the computation of attention accuracy (Section 4.4), and presents additional experimental results.

## A.1. Interpretations

### A.1.1. Invariant Loss is not Intervention?

Given the causal intervention formulation (Eq. (1)), readers may consider that the implementation of backdoor adjustment in robust classification can be simply achieved by optimizing the cross entropy loss in each data split $t \in \mathcal{T}$, rather than the invariant loss. Please note that this claim is actually to use the first objective item of the invariant loss and it acts just like the conventional cross entropy. The reason is, the backdoor adjustment in statistical causality theory is not designed for learning process in computer vision practice. The sum "$\sum$" of backdoor adjustment can not be implemented by summing up the cross entropy loss directly. We need the second term in Eq. (5) as a regularization to effectively implement the "$\sum$" by collecting the common invariant representation across different splits $t$. Then during inference, we just discard the second regularization term and forward the model to get the intervened prediction.

### A.1.2. Mediator in Causal Graph

We have introduced the mediator $M$ in the main paper and it is a part of the causal effect which need to be retained. In this paper we manually separate it out to better illustrate the improper intervention and explain why it hurts the causal effects. That is, non-accurate confounder set which falsely contains $M$ will lead to the over adjustment of the mediator and hurt the causal representation. Then in Section 3.2 and 3.3 of the main paper, we give details of our proposed CaaM for obtaining better confounder set.

## A.2. Theoretical Proofs

### A.2.1. Proof of Improper Causal Intervention

We will show the derivation for the backdoor adjustment formula using the three rules of *do*-calculus [14], whose detailed proof can be found in [14, 13]. For a causal directed acyclic graph $\mathcal{G}$, let $X, Y, Z$ and $W$ be arbitrary disjoint sets of nodes. We use $\mathcal{G}_{\overline{X}}$ to denote the manipulated graph where all incoming arrows to node $X$ are deleted. Similarly $\mathcal{G}_{\underline{X}}$ represents the graph where outgoing arrows from node $X$ are deleted. We use lower case $x, y, z$ and $w$ for specific values taken by each set of nodes: $X = x, Y = y, Z = z$ and $W = w$. For any interventional distribution compatible with $\mathcal{G}$, we have the following three rules:

**Rule 1** Insertion/deletion of observations. If $(Y \perp\!\!\!\perp Z | X, W)_{\mathcal{G}_{\overline{X}}}$:

$$P(y|do(x), z, w) = P(y|do(x), w), \quad (A1)$$

**Rule 2** Action/observation exchange. If $(Y \perp\!\!\!\perp Z | X, W)_{\mathcal{G}_{\overline{X}\underline{Z}}}$,

$$P(y|do(x), do(z), w) = P(y|do(x), z, w), \quad (A2)$$

**Rule 3** Insertion/deletion of actions. If $(Y \perp\!\!\!\perp Z | X, W)_{\mathcal{G}_{\overline{X}\overline{Z(W)}}}$,

$$P(y|do(x), do(z), w) = P(y|do(x), w), \quad (A3)$$

where $Z(W)$ is the set of nodes in $Z$ that are not ancestors of any $W$-node in $\mathcal{G}_{\overline{X}}$.

In our causal graph, the desired interventional distribution $P(Y|do(X))$ can be derived by:

$$P(Y|do(X)) \quad (A4)$$

$$= \sum_s P(Y|do(X), S=s) P(S=s|do(X)) \quad (A5)$$

$$= \sum_s P(Y|do(X), S=s) P(S=s) \quad (A6)$$

$$= \sum_s P(Y|X, S=s) P(S=s), \quad (A7)$$

where Eq. (A5) follows the law of total probability; Eq. (A6) uses Rule 3 given $S \perp\!\!\!\perp X$ in $\mathcal{G}_{\overline{X}}$; Eq. (A7) uses Rule 2 to change the intervention term to observation as $(Y \perp\!\!\!\perp X | S)$ in $\mathcal{G}_{\underline{X}}$. S When $M$ and $S$ are *disentangled* (or conditional independent), *i.e.*, $(S \perp\!\!\!\perp M)|X$, the backdoor adjustment formula can be further written as Eq. (2) by

$$P(Y|do(X)) \quad (A8)$$

$$= \sum_s P(Y|X, S=s) P(S=s), \quad (A9)$$

$$= \sum_s \sum_m P(Y|X, m, s) P(m|X, s) P(s), \quad (A10)$$

$$= \sum_s \sum_m P(Y|X, m, s) P(m|X) P(s), \quad (A11)$$

where Eq. (A10) follows the law of total probability and Eq. (A11) is due to the conditional independence between $M$ and $S$ given $X$. This proves Eq. (2). However when $M$ and $S$ are *entangled* (as the "improper intervention"), $P(m|X,s) \neq P(m|X)$ and Eq. (A10) $\neq$ Eq. (A11). Therefore, in the case of entanglement, the optimization objective becomes Eq. (3).

### A.2.2. Proof of Convergence of CaaM

In this section, we will first give the detailed intuition for the effectiveness of our CaaM adversarial training. Then we prove the positive feedback between the Maxi-Game and Mini-Game and the existence of the global optimal point of our CaaM, which ensures the convergence of our algorithm.

**Intuitive Explanation.** Consider the whole pipeline of the robust prediction. First, we need the complementary attention to disentangle the causal and confounder feature (*i.e.*, $c$ and $s$) from the image representation $\mathbf{x}$ given a data partition $\mathcal{T}_i$; then we use the confounder feature $s$ to generate better partition $\mathcal{T}_{i+1}$. The key is the mechanism for $c$ and $s$ to mutual promote each other and there exists a positive feedback between these 2 steps. With an accumulative confounder set, more accurate invariant representation $c$ can be encoded, which will further bring a clearer picture of confounder feature $s$. Therefore we can then promote the partition $\mathcal{T}$ with the better $s$.

**Positive Feedback.** Here we assume that the image representation $\mathbf{x}$ is made up of the causal feature $c$ and confounder feature $s$: $\mathbf{x} = c \circ s$, where $\circ$ is feature fusion. Here we use $c^*$ and $s^*$ to represent the oracle causal and confounder representation what we tend to find. $\mathcal{T}^*$ and $\theta^*$ are the corresponding oracle data partition. The initialized bias model is named as $\Omega$ which is trained with the conventional cross entropy loss. Therefore, we have $\Omega(x) = \mathbf{x}$ and it can be regarded as training with the random data partition. With Eq. (7), we have:

$$\theta_0 = \max_{\theta} \text{IL}(h, \Omega(x), \mathcal{T}(\theta)). \quad (A12)$$

$\theta_0$ and its counterpoint $\mathcal{T}_0$ are the better partition compared to the random $\theta$, approaching to $\mathcal{T}^*$ and $\theta^*$. The behind reason can also be explained with the heterogeneous environment theorem [12]:

**Theorem A1.** *Denote image $X$ and label $Y$, using the functional representation lemma [6], there exists random variable $s$ such that $X = X(c,s)$ to make $P_t(Y|s)$ arbitrarily change across data splits $t$.*

Actually $c$ and $s$ are the robust causal and non-robust context feature. Theorem A1 indicates that a good data split $t$ should reveal as much as possible spurious (or variant) feature to help to narrow the invariant feature. That means, if we can access more accurate bias feature, we can then achieve the better data split with Maxi-Game. Since the initial model $\Omega$ captures part of bias feature, the optimized $\theta_0$ ($\mathcal{T}_0$) is better than random partition. Please note that the Maxi-Game only takes the confounder feature in $\overline{\mathcal{A}}(x)$ to generate the data partition as the causal feature does not contribute to the maximization (causal feature is always invariant across all distributions).

Then consider Mini-Game using Eq. (6), we can disentangle causal feature $c_1$ and confounder feature $s_1$ under current $\mathcal{T}_0$ with the proposed complementary attention module:

$$c_1, s_1 = \min_{\mathcal{A}_1, \overline{\mathcal{A}}_1, f, g} \text{XE}(f, \widetilde{x}, \mathcal{D}) + \text{IL}(g, \mathcal{A}_1(x), \mathcal{T}_0). \quad (A13)$$

Eq. (A13) directly optimize $c$ and $\widetilde{x}$. Similarly, note that minimizing the IL loss will discard the confounder feature rather than hurt the causal feature, as the causal feature is always invariant across all distributions. With our assumption that $\mathbf{x} = c \circ s \approx \widetilde{x}$, better $c$ and $\widetilde{x}$ leads to more accurate $s$. Therefore, confounder feature $s_1$ is a better approximation to $s^*$ than previous feature $\mathbf{x}$ and the fitted bias model $h_1(\overline{\mathcal{A}}_1(x))$ is better than $\Omega$ using:

$$h_1 = \min_h \text{XE}(h, \overline{\mathcal{A}}_1(x), \mathcal{D}). \quad (A14)$$

We next update the new partition $\theta_1$ and $\mathcal{T}_1$ with Eq. (A12). Since the bias model is better, the obtained $\theta_1$ and $\mathcal{T}_1$ will be closer to $\mathcal{T}^*$ and $\theta^*$. Moreover, since CaaM does not introduce new image label space (*e.g.*, continual learning [19]) but just update the partition, the previous $\mathcal{T}$ will still play its role partially. This results in an even better intervention with a comprehensive confounder set. Until now, we have illustrated a complete positive feedback of our CaaM. Next we give that the Mini-Game and Maxi-Game can both reach the global optimal point.

**Global Optimal Point.** Given the partition $\mathcal{T} = \mathcal{T}^*$, the learned $c$ will be the oracle causal representation $c^*$ when Eq. (A13) achieves the global minimal point. Then under the assumption of $\mathbf{x} = c \circ s$, $s$ is equal to $s^*$. Therefore, the corresponding $\mathcal{T}$ can not be better. That is, Eq. (A12) also arrives to the global optimal point.

## A.3. Implementation Details

### A.3.1. Details of Invariant Loss

While having represented the core function of Invariant Loss (Eq. (5)) in the main paper, we aim to present more details regarding its motivation and practical implementation in this section. Invariant Loss aims to find an *invariant representation* (*i.e.*, the causal feature) robust for prediction, such that the optimal classifier over the representation is the same across dataset partition $\mathcal{T}$ [2]. This is formally given by:

**Definition A1** (Invariant Representation). *A representation $\mathcal{A}(x) \in \mathbb{R}^d$ is invariant across partition $\mathcal{T}$ if there exists a classifier $g : \mathbb{R}^d \to \mathbb{R}^k$ such that for $\forall t \in \mathcal{T}$, $g \in \arg\min_{\bar{g}} \text{XE}(\bar{g}, \mathcal{A}(x), t)$.*

This is achieved by the following objective function:

$$\min_{\mathcal{A},g} \sum_{t \in \mathcal{T}} \text{XE}(g, \mathcal{A}(x), t) \tag{A15}$$
$$\text{s.t.} g \in \arg\min_{\bar{g}} \text{XE}(\bar{g}, \mathcal{A}(x), t), \forall t \in \mathcal{T},$$

where the definition of XE is given in Section 3.2. Intuitively, this objective optimizes the empirical risk minimization (ERM) representation $\mathcal{A}(x)$ subject to an invariant representation. However, this is a complex bi-level optimization problem, where each constraint corresponds to an inner-loop optimization. Therefore, [2] proposes a practical objective for approximation given in Eq. (5), *i.e.*,

$$\min_{\mathcal{A},g} \sum_{t \in \mathcal{T}} \text{XE}(g, \mathcal{A}(x), t) + \lambda \|\nabla_{\mathbf{w}=\mathbf{1.0}} \text{XE}(\mathbf{w}, \mathcal{A}(x), t)\|_2^2. \tag{A16}$$

In this paper, we exactly use Eq. (A16) for the Invariant Loss in Eq. (7), but for that in Eq. (6), we follow a more practical version [16] of Eq. (A16). Specifically, different linear classifier $\mathbf{W}_t$ is initialized for each data split $t$. The causal feature should be stable across splits. Therefore, the classifiers $\mathbf{W}_t$ are encouraged to converge to a common matrix, leading to a more practical version to replace the gradient penalty of Eq. (A16):

$$\min_{\mathcal{A},\{\mathbf{W}_t\}_{t=1}^m} \sum_{t \in \mathcal{T}} \text{XE}(\mathbf{W}_t, \mathcal{A}(x), t) + \lambda \underset{t}{\text{Var}}(\mathbf{W}_t), \tag{A17}$$

where $\underset{t}{\text{Var}}(\mathbf{W}_t)$ controls the variance of classifier weights:

$$\underset{t}{\text{Var}}(\mathbf{W}_t) = (1/m) \sum_t (\|\mathbf{W}_t - \overline{\mathbf{W}}\|_2 / \|\mathbf{W}_t\|_1)^2, \tag{A18}$$

where $\overline{\mathbf{W}}$ is the arithmetic mean of $\mathbf{W}_t$ over $t$. During testing, we use $\overline{\mathbf{W}}$ as classifier $g$ to make the causal prediction $g(\mathcal{A}(x))$. Based on this invariant loss, we futher design a novel complementary attention to disentangle causal representation and adversarial training pipeline to update data partition.

### A.3.2. Details of Adversarial Training

**Mini-Game.** The Mini-Game (Eq. (6)) can be further divided into 2 sub-steps. The first is the intervention training:

$$\min_{\mathcal{A},\overline{\mathcal{A}},f,g} \text{XE}(f, \widetilde{x}, \mathcal{D}) + \text{IL}(g, \mathcal{A}(x), \mathcal{T}). \tag{A19}$$

Then after achiving the confounder feature $\overline{\mathcal{A}}(x)$, we explicitly trains a biased classifier $h$ using the confounding feature $\overline{\mathcal{A}}(x)$ and the original label $y$. To achieve this, we fix $\overline{\mathcal{A}}$ and optimize $h$ by minimizing the CE loss:

$$\min_h \text{XE}(h, \overline{\mathcal{A}}(x), \mathcal{D}). \tag{A20}$$

The Eq. (A19) and Eq. (A20) constitute the Mini-Game.

**Maxi-Game.** The core of Maxi-Game is to optimize Eq. (7), *i.e.*, maximizing the IL loss to update the partition $\mathcal{T}_i$:

$$\max_\theta \ \text{IL}(h, \overline{\mathcal{A}}(x), \mathcal{T}_i(\theta)). \tag{A21}$$

We have introduced in the main paper that we define a set of optimizable parameters $\theta \in \mathbb{R}^{K \times m}$. Each parameter $\theta_{p,q}$ stores the current probability of the $p$-th sample belonging to the $q$-th split ($t_q \in \mathcal{T}_i$). $\mathcal{T}_i(\theta)$ is obtained by assign each data sample to the split index with largest probability. However, this process contains the `argmax` function which is not continuous in the backward pass. In this paper we use the Gumbel-Softmax [10] trick to relax the discrete sampling operation. Specifically, consider a $k$-dimensional categorical probabilities $\pi_1, ..., \pi_k$. The sample $\mathbf{y} = (y_1, ..., y_k)$ is given by:

$$y_v = \frac{\exp\left((\log(\pi_v) + \mu_v)/\tau\right)}{\sum_{j=1}^k \exp\left((\log(\pi_j) + \mu_j)/\tau\right)} \tag{A22}$$

where $\tau = 1.0$ is a temperature parameter. $\mu_v = -\log(-\log(u_v))$ and $u_v \sim \text{Uniform}(0, 1)$. Here $\mu_v$ is named *Gumbel Noise*, perturbs each $\log(\pi_v)$ term so that taking the original `argmax` becomes equivalent to drawing a sample from the distribution $\pi_1, ..., \pi_k$. Moreover, to facilitate dataset grouping, the hard sample is needed. Therefore, we further adopt the Straight-Through (ST) Gumbel-Softmax [4, 10]. In the forward pass, it discretizes a continuous probability vector $\mathbf{y}$ sampled from the Gumbel-Softmax distribution into the one-hot vector $\mathbf{y}^{ST}$, where:

$$y_v^{ST} = \begin{cases} 1 & v = \arg\max_j y_j \\ 0 & \text{otherwise} \end{cases} \tag{A23}$$

And in the backward pass it simply uses the continuous $\mathbf{y}$, thus the loss signal is still able to backpropagate.

**Initialization.** For initialization, we train a bias model $h, \overline{\mathcal{A}}$ with the conventional cross entropy loss and assume this model is overfitted to some extent of confounding effects. Then we can get the initial $\mathcal{T}_0$ by replacing $h, \overline{\mathcal{A}}$ with $h_0, \overline{\mathcal{A}}_0$ in the Maxi-Game Eq. (A21). That means, we initialize the confounder representation with $h_0(\overline{\mathcal{A}}_0(x))$.

The overall training pipeline is summarized in Algorithm 1.

## A.4. Experimental Details

### A.4.1. NICO Dataset

In our experiment, we selected a subset of NICO animal dataset [9] as a challenging benchmark to test OOD robustness for proposed CaaM and baselines. Specifically, images in NICO are labeled with a context class (*e.g.*, "on grass"), besides the object class (*e.g.*, "dog"). As discussed in section 4.1, during training we chose 7 context classes

| Class \ Context | Long-Tailed Contexts | | | | | | | Zero-shot Contexts | | |
|---|---|---|---|---|---|---|---|---|---|---|
| Dog | on grass | in water | in cage | eating | on beach | lying | running | at home | in street | on snow |
| Cat | on snow | at home | in street | walking | in river | in cage | eating | in water | on grass | on tree |
| Bear | in forest | black | brown | eating grass | in water | lying | on snow | on ground | on tree | white |
| Sheep | eating | on road | walking | on snow | on grass | lying | in forest | aside people | in water | at sunset |
| Bird | on ground | in hand | on branch | flying | eating | on grass | standing | in water | in cage | on shoulder |
| Rat | at home | in hole | in cage | in forest | in water | on grass | eating | lying | on snow | running |
| Horse | on beach | aside people | running | lying | on grass | on snow | in forest | at home | in river | in street |
| Elephant | in zoo | in circus | in forest | in river | eating | standing | on grass | in street | lying | on snow |
| Cow | in river | lying | standing | eating | in forest | on grass | on snow | at home | aside people | spotted |
| Monkey | sitting | walking | in water | on snow | in forest | eating | on grass | in cage | on beach | climbing |

Table A1. Construction of our NICO [9] subset for OOD multi-classification. **Context** denotes the context class name, while **Class** represents the object class name. "Long-Tailed Contexts" is the training contexts arranged by the sample number order (from more to less) and "Zero-shot Contexts" represents the context labels only appear in testing rather than training.

---

**Algorithm 1** Adversarial CaaM Training

1: **Input:** Dataset $\mathcal{D}$
2: **Output:** $\mathcal{A}, \overline{\mathcal{A}}, \theta$
3: Initialize split assignment $\theta_0, \mathcal{T}_0$ with Eq. (A12)
4: Initialize $f, g, h, \mathcal{A}, \overline{\mathcal{A}}$
5: **for** $i \in \{1, 2, ..., N\}$ **do**
6:    **for each** $x \in \mathcal{D}$ **do**
7:       $\mathbf{c}_i \leftarrow \mathcal{A}_i(x)$
8:       $\mathbf{s}_i \leftarrow \overline{\mathcal{A}}_i(x)$
9:    **end for**
10:   Update $f_i, g_i, h_i, \mathcal{A}_i, \overline{\mathcal{A}}_i$ with Eq. (A19), Eq. (A20) and $\mathcal{T}_i$;     {// Mini-Game}
11:   Update $\theta_i, \mathcal{T}_i$ with Eq. (A21);     {// Maxi-Game}
12: **until** end

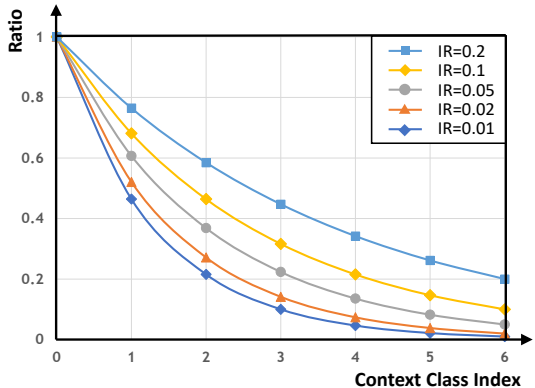

Figure A1. Plot of context class index against its corresponding ratio under various imbalance ratio (IR).

(Long-Tailed Contexts as shown in Table A1) for each object class. Next, we formed a long-tailed training dataset by selecting part of the images in each context class with multiplying a ratio. In particular, the ratio for $w$-th context class ($w \in \{0, \ldots, 6\}$) is given by

$$\text{ratio} = \text{IR}^{w/6}, \tag{A24}$$

where IR is a hyper-parameter that denotes the imbalance ratio. The effect of IR on ratio is shown in Figure A1 — lower ratio leads to the harder OOD problem. In the main paper we keep IR = 0.02. During testing, the number of test samples across the 7 context classes is balanced, *i.e.*, 50 samples per context. Moreover, we added 3 zero-shot context classes for each object class as shown in Table A1 (last three columns). These zero-shot context classes have the larger number of test samples (100 samples per context). Therefore, a model that performs well in our split must be robust to both long-tailed and zero-shot problems *w.r.t.* the context class. Figure A2 shows an example of our constructed subset for "cat" and "dog" during training and testing. Moreover, to construct $m$ ground truth splits on NICO dataset for training in "w/ H.M. $\mathcal{T}$" setting, we first sort images in the context order and then divide them into $m$ equal parts.

### A.4.2. ImageNet-9 Dataset

Specifically, we elaborate the construction of the proxy context labels, respectively, for ImageNet-9 and ImageNet-A.

**Proxy Context Label for ImageNet-9.** Note that there is no ground truth context annotation in ImageNet-9 training and testing set. Therefore, we follow [3] to obtain the proxy ground truth context labels using texture feature clustering. Specifically, we extract the texture features from images by computing the gram matrices of low-layer feature maps [7, 11] from `relu1_2` of the ImageNet pre-trained VGG16 [15]. Then we run the mini-batch $k$-means algorithm with $k = m$ with batch size 1024, $m$ is the number of data splits. For the construction of the unbias test set, we follow [3] to set $k = 9$ with batch size 1024. The clustering examples are shown in Figure A3 (left).

**ImageNet-A.** ImageNet-A is a dataset of natural adversarial examples for ImageNet classifiers, or real-world examples that fool current classifiers as shown in Figure A3 (right). The images consist of many failure modes of net-

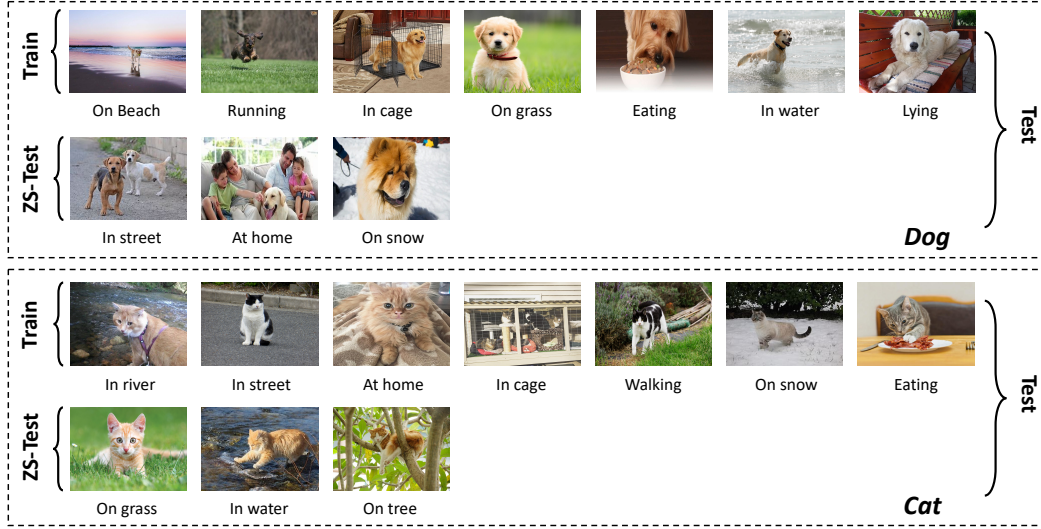

Figure A2. We list the sample images of each context class using "Dog" and "Cat" as the example in our constructed NICO dataset. **Train**, **Test** and **ZS-Test** denote samples for training, testing and zero shot testing respectively. Note that there is no overlap between training and testing images.

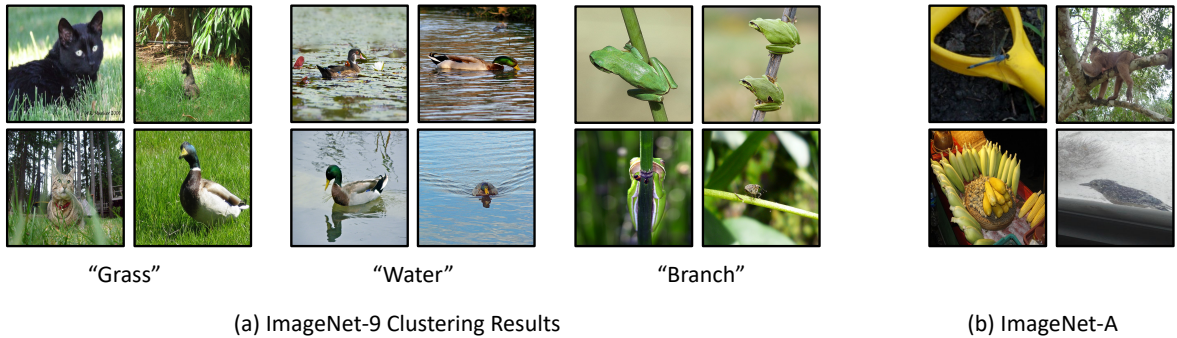

(a) ImageNet-9 Clustering Results

(b) ImageNet-A

Figure A3. The examples of the clustering results of ImageNet-9 dataset and the samples of ImageNet-A dataset.

works when "frequently appearing background elements" become erroneous cues for recognition (*e.g.*, a dragonfly on a yellow metal bracket is recognised as the banana).

### A.4.3. Training Details

In Mini-Game, the optimizer was set to SGD with a learning rate of 0.05 for ResNet model; while for T2T-ViT, the learning rate was set as 0.001 with AdamW optimizer following ViT [5]. We trained the model with 200 epochs for NICO dataset and the learning rate was decreased by 5 at 80, 120, 160 epoch. While for ImageNet-9 we trained for 120 epochs with learning rate decreased by 5 at 50, 80, 100 epoch. For the bias classifier fitting Eq. (A20) and the Maxi-Game Eq. (A21), we trained each for 100 epochs with early stopping (accuracy no longer increases more than 5 epoch). The optimizer was set to SGD with learning rate as 0.1. For Mini-Game, $\lambda$ in invariant loss was set to 5e4 or 5e5 following previous paper [16]. In Maxi-Game, $\lambda$ is fixed to 1e6. $N$ was set from 5 to 20. In Table (2), for results in first two row (**Num L.** and **Num S.**), we did not use the adversarial training. The default $M$ was 2 for CNN-CaaM and 4 for ViT-CaaM, while the default $m$ was set to 4.

### A.4.4. Attention Accuracy (Q3)

In the question 3 of the main paper, we calculate the attention accuracy for proposed method and its comparisons to quantify the robustness of CaaM attention on ImageNet dataset. Here we detail how to compute. Given an Image, we can obtain its attention map with different architectures. Specifically, for ResNet+CBAM, we take the CAM activation [20] as the attention map following the original CBAM paper [17]. While for T2T-ViT, we use the Attention Rollout [1] following original ViT [5]. Briefly, we averaged attention weights of T2T-ViT across all heads and then re-

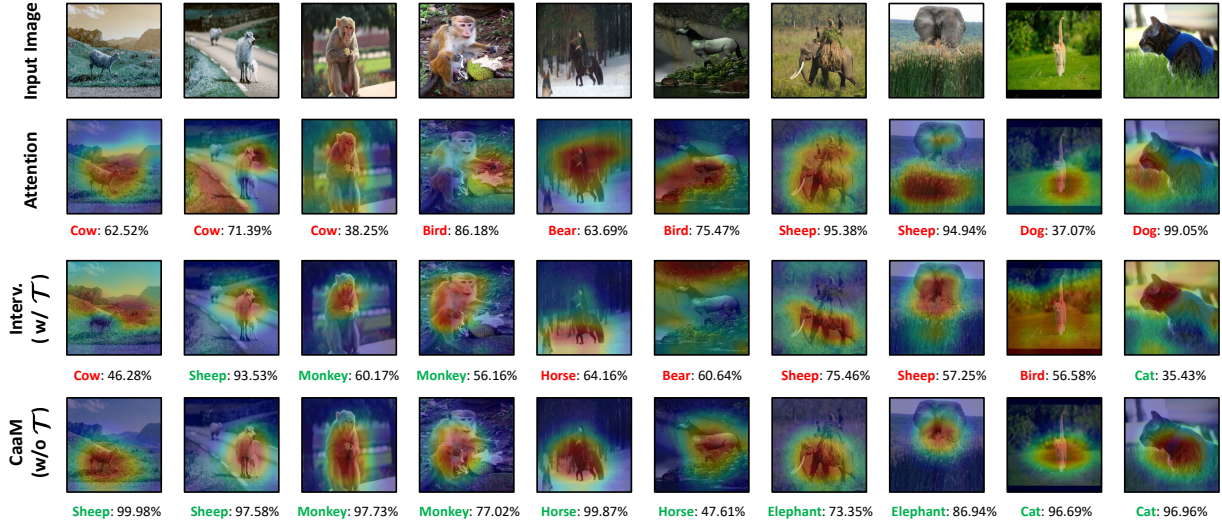

Figure A4. The attention activation maps based on CNN on NICO dataset. "Attention" and "Interv." denote the conventional attention model and intervention method [16] respectively. **Red** and **Green** text denote the false and correct prediction followed by the prediction confidence.

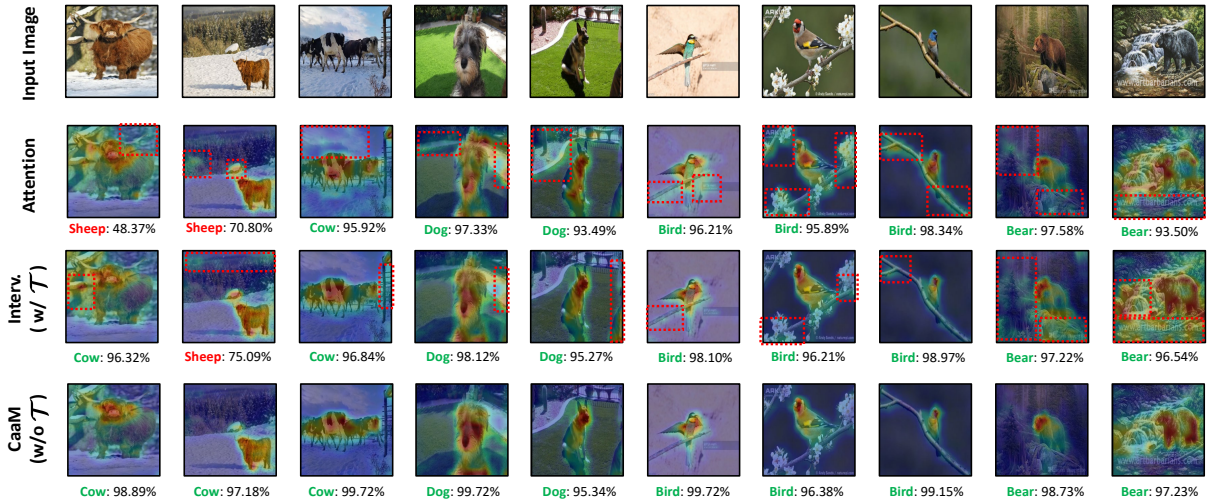

Figure A5. The attention activation maps based on T2T-ViT on NICO dataset. "Attention" and "Interv." denote the conventional attention model and intervention method [16] respectively. **Red** and **Green** text denote the false and correct prediction followed by the prediction confidence. The red dashed boxes highlight the worse attention activation for the comparisons.

cursively multiplied the weight matrices of all layers. This accounts for the mixing of attention across tokens through all layers. Having the ground truth object location bounding box annotation $B$ and the attention map $A$ which is a weight matrix of image size, we can then compute the attention accuracy by:

$$\text{Att. Acc.} = \frac{sum((A \cap B) > \sigma)}{sum(A > \sigma)}, \quad (A25)$$

where $A \cap B$ denotes the attention map area in the bounding box and $\sigma = 0.9$ is the threshold. This function is similar to the Intersection over Union (IoU) score metric in object detection.

### A.4.5. Additional Results

**Complexities.** We show the model sizes and the computational costs in Table A2. We can see that compared to baseline models ResNet18+CBAM and T2T-ViT7, using CaaM adds a small number of network parameters (overhead). This is because the complementary attention in CaaM does not rely on new network layers. In terms of computing speed, single-layer CaaM has the comparable Flops and

| Models | Params (M) | Flops (G) | MACs (G) |
|---|---|---|---|
| ResNet18 [8] | 11.18 | 3.63 | 1.81 |
| ResNet18+CBAM [17] | 11.27 | 3.64 | 1.82 |
| ResNet18+CaaM | 11.29 | 3.64 | 1.82 |
| ResNet18+CaaM ($M$=2) | 11.29 | 4.46 | 2.23 |
| ResNet18+CaaM ($M$=4) | 11.29 | 5.28 | 2.64 |
| T2T-ViT7 | 4.00 | 1.95 | 0.97 |
| T2T-ViT7+CaaM | 4.01 | 2.08 | 1.04 |
| T2T-ViT7+CaaM ($M$=2) | 4.01 | 2.20 | 1.10 |
| T2T-ViT7+CaaM ($M$=4) | 4.01 | 2.46 | 1.23 |

Table A2. The model size and computational cost comparison between our proposed CaaM and baseline models.

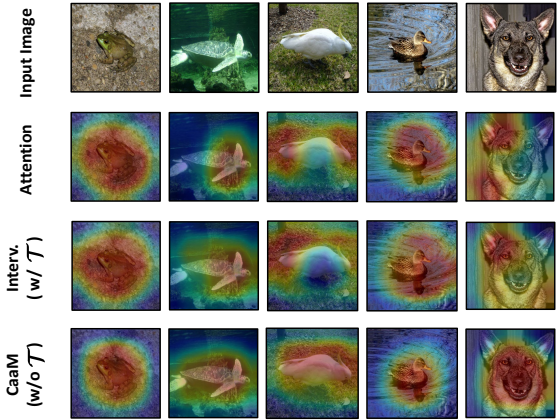

Figure A6. The attention activation maps based on CNN on ImageNet-9 dataset. "Attention" and "Interv." denote the conventional attention model and intervention method [16] respectively.

MACs to baseline models. Multi-layer CaaM has linearly-increased Flops and MACs with respect to the number of layers, while the maximum costs are tolerable.

**Visualizations.** In Figure A4 and Figure A5, we supplement more visualization results for several comparable methods: our CaaM (unsupervised, *i.e.*, without using the labels of partition $\mathcal{T}$), an intervention method [16] (supervised, *i.e.*, using $\mathcal{T}$), and a convention attention method (CBAM [17] for CNN and T2T-ViT [18] for ViT). Samples in Figure A4 are the results of CNN-based models. Samples in Figure A5 are from ViT-based models. Both are from the NICO dataset. For each sample, we report the heatmap of attention, the predicted label and the corresponding probability.

In Figure A4, we can observe that 1) the conventional attention model (the second row) produces many inaccurate attentions on images and false predictions of object labels; 2) the intervention method (the third row) partially tackle the problems by using the labeled partitions $\mathcal{T}$; and 3) impressively, our CaaM (the forth row) achieves both more accurate predictions and more precise attentions—mostly focused on the object bodies. Similar results can also be drawn from Figure A5 for methods based on ViT.

In addition to the quantitative attention accuracy results on ImageNet-9 dataset (Table 3) in the main paper, here we also visualize the attention map based on CNN on ImageNet-9 biased validation set in Figure A6, denoting the *IID* setting (NICO, ImageNet-9 unbiased set and ImageNet-A are the *OOD* setting). Compared to convention attention and intervention method, we can see that our CaaM can still clearly get tighter and more explainable attention maps in IID setting.

# References

[1] Samira Abnar and Willem Zuidema. Quantifying attention flow in transformers. *arXiv preprint arXiv:2005.00928*, 2020. 5

[2] Martin Arjovsky, Léon Bottou, Ishaan Gulrajani, and David Lopez-Paz. Invariant risk minimization. *arXiv preprint*, 2019. 2, 3

[3] Hyojin Bahng, Sanghyuk Chun, Sangdoo Yun, Jaegul Choo, and Seong Joon Oh. Learning de-biased representations with biased representations. In *ICML*, pages 528–539. PMLR, 2020. 4

[4] Yoshua Bengio, Nicholas Léonard, and Aaron Courville. Estimating or propagating gradients through stochastic neurons for conditional computation. *arXiv preprint arXiv:1308.3432*, 2013. 3

[5] Alexey Dosovitskiy, Lucas Beyer, Alexander Kolesnikov, Dirk Weissenborn, Xiaohua Zhai, Thomas Unterthiner, Mostafa Dehghani, Matthias Minderer, Georg Heigold, Sylvain Gelly, et al. An image is worth 16x16 words: Transformers for image recognition at scale. *arXiv preprint*, 2020. 5

[6] Abbas El Gamal and Young-Han Kim. *Network information theory*. Cambridge university press, 2011. 2

[7] Leon A Gatys, Alexander S Ecker, and Matthias Bethge. Texture synthesis using convolutional neural networks. *arXiv preprint arXiv:1505.07376*, 2015. 4

[8] Kaiming He, Xiangyu Zhang, Shaoqing Ren, and Jian Sun. Deep residual learning for image recognition. In *CVPR*, pages 770–778, 2016. 7

[9] Yue He, Zheyan Shen, and Peng Cui. Towards non-iid image classification: A dataset and baselines. *Pattern Recognit.*, 110:107383, 2021. 3, 4

[10] Eric Jang, Shixiang Gu, and Ben Poole. Categorical reparameterization with gumbel-softmax. *arXiv preprint arXiv:1611.01144*, 2016. 3

[11] Justin Johnson, Alexandre Alahi, and Li Fei-Fei. Perceptual losses for real-time style transfer and super-resolution. In *European conference on computer vision*, pages 694–711. Springer, 2016. 4

[12] Jiashuo Liu, Zheyuan Hu, Peng Cui, Bo Li, and Zheyan Shen. Heterogeneous risk minimization. *arXiv preprint arXiv:2105.03818*, 2021. 2

[13] Judea Pearl. Causal diagrams for empirical research. *Biometrika*, 82(4):669–688, 1995. 1

[14] Judea Pearl. *Causality*. Cambridge university press, 2009. 1

[15] Karen Simonyan and Andrew Zisserman. Very deep convolutional networks for large-scale image recognition. *arXiv preprint arXiv:1409.1556*, 2014. 4

[16] Damien Teney, Ehsan Abbasnejad, and Anton van den Hengel. Unshuffling data for improved generalization. *arXiv preprint*, 2020. 3, 5, 6, 7

[17] Sanghyun Woo, Jongchan Park, Joon-Young Lee, and In So Kweon. Cbam: Convolutional block attention module. In *ECCV*, pages 3–19, 2018. 5, 7

[18] Li Yuan, Yunpeng Chen, Tao Wang, Weihao Yu, Yujun Shi, Francis EH Tay, Jiashi Feng, and Shuicheng Yan. Tokens-to-token vit: Training vision transformers from scratch on imagenet. *CVPR*, 2021. 7

[19] Friedemann Zenke, Ben Poole, and Surya Ganguli. Continual learning through synaptic intelligence. In *International Conference on Machine Learning*, pages 3987–3995. PMLR, 2017. 2

[20] Bolei Zhou, Aditya Khosla, Agata Lapedriza, Aude Oliva, and Antonio Torralba. Learning deep features for discriminative localization. In *Computer Vision and Pattern Recognition*, 2016. 5



\end{document}